%% file: paper.tex
\documentclass[conference]{IEEEtran}
\IEEEoverridecommandlockouts
\usepackage{cite}
\usepackage{amsmath,amssymb,amsfonts}
\usepackage{graphicx}
\usepackage{textcomp}
\usepackage{xcolor}
\def\BibTeX{{\rm B\kern-.05em{\sc i\kern-.025em b}\kern-.08em
    T\kern-.1667em\lower.7ex\hbox{E}\kern-.125emX}}

\usepackage{algorithm}
\usepackage[noend]{algpseudocode}

\begin{document}

\title{Evolutionary Approach to Collectible Card Game Arena Deckbuilding using Active Genes
\thanks{This work was supported by the National Science Centre, Poland under project number 2017/25/B/ST6/01920.} 
}

\author{\IEEEauthorblockN{Jakub Kowalski}
\IEEEauthorblockA{\textit{Institute of Computer Science} \\
\textit{University of Wroc{\l}aw}\\
Wroc{\l}aw, Poland \\
jko@cs.uni.wroc.pl}
\and
\IEEEauthorblockN{Rados{\l}aw Miernik}
\IEEEauthorblockA{\textit{Institute of Computer Science} \\
\textit{University of Wroc{\l}aw}\\
Wroc{\l}aw, Poland \\
radekmie@gmail.com}
}

\maketitle

\begin{abstract}

In this paper, we evolve a card-choice strategy for the arena mode of Legends of Code and Magic, a programming game inspired by popular collectible card games like Hearthstone or TES: Legends. In the arena game mode, before each match, a player has to construct his deck choosing cards one by one from the previously unknown options.
Such a scenario is difficult from the optimization point of view, as not only the fitness function is non-deterministic, but its value, even for a given problem instance, is impossible to be calculated directly and can only be estimated with simulation-based approaches.  

We propose a variant of the evolutionary algorithm that uses a concept of an \emph{active gene} to reduce the range of the operators only to generation-specific subsequences of the genotype. Thus, we batched learning process and constrained evolutionary updates only to the cards relevant for the particular draft, without forgetting the knowledge from the previous tests.

We developed and tested various implementations of this idea, investigating their performance by taking into account the computational cost of each variant. Performed experiments show that some of the introduced active-genes algorithms tend to learn faster and produce statistically better draft policies than the compared methods.

\end{abstract}



\section{Introduction}

Currently, not only classical boardgames like Chess \cite{Campbell2002Deep} and Go \cite{Silver2016Mastering} are used as grand challenges for AI research.
It has been recently shown that such a role may be taken by modern computer games.
So far presented approaches that beat the best human players in \emph{Dota~2} \cite{OpenAIDota} and \emph{StarCraft II} \cite{vinyals2019grandmaster} are one of the most spectacular and media-impacting demonstrations of AI capabilities.

The weight is put on particular game features that make designing successful AI players especially tricky, e.g., imperfect information, randomness, long term planning, and massive action space. One of the game genres containing all these game features is Collectible Card Games \cite{hoover2019many}. 

Recently, numerous research has been conducted in this domain, focusing mainly on development of MCTS-based agents, and creating the deck recommendation systems that will choose the right set of cards to play. 
The \emph{Hearthstone AI Competition} \cite{HearthstoneAICompetition}, with the goal to develop the best agent for the game \emph{Hearthstone}, \cite{Blizzard2004Hearthstone} was organized during the IEEE CIG/COG conferences in 2018 and 2019. 
The \emph{Strategy Card Game AI Competition} based on programming game \emph{Legends of Code and Magic} (LOCM) \cite{LOCMPage}, designed especially for handling AI vs.\ AI matches and played in the arena mode, was hosted in 2019 by IEEE CEC and IEEE COG.
The \emph{AAIA’17 Data Mining Challenge: Helping AI to Play Hearthstone} \cite{janusz2017helping} was focused on developing a scoring model for predicting win chances of a player, based on single-game state data.

This work is the first approach to build a deck recommendation system for the \emph{arena} mode. As in the other modes players can use their full collection of cards, in arena they draw a deck from a random selection of cards before every game. Such a task is characterized by an even larger domain (considering all possibilities of available choices), higher non-determinism (the additional card selection phase is non-deterministic), and harder opponent prediction.

We propose variants of the evolutionary algorithm that uses a concept of an \emph{active gene} to reduce the range of the operators only to specific subsequences of the genotype that changes from generation to generation. In other words, our batched learning constrains evolutionary updates only to the cards relevant for the particular draft.
Individuals forming the next population are no longer simply selected among the parents/offspring populations, but instead, they are merged from the specific subsets of their representatives.

We have conducted a series of experiments in LOCM to estimate the performance of multiple variants of designed algorithms and compare them with several baselines.
Algorithms learn on a small sample of random drafts (available card choices) and are tested on a larger number of other drafts, using large number of game simulations to estimate the fitness value in such a highly non-deterministic environment.
The results show that some of the introduced active genes variants perform better than other tested approaches, i.e., they tend to draw stronger decks from the same available card choices.


\section{Background}

\subsection{Collectible Card Games}

\emph{Collectible Card Game} (CCG) is a broad genre of both board and digital games. Starting with \emph{Magic: The Gathering} (MtG) \cite{WotC1993MtG} in the early 90s, or more recent \emph{Hearthstone} \cite{Blizzard2004Hearthstone} and \emph{The Elder Scrolls: Legends} (TESL) \cite{Bethesda2017TESL}, a huge number of similar games have been created.

The mechanics differ between games, but basics are similar. Two players with their \emph{decks} draw an initial set of cards into their \emph{hands}. Then the actual play begins. A single \emph{turn} consists of a few \emph{actions}, like playing a card or using an onboard card. The game ends as soon as one of the players wins, most often by getting their opponent's health to zero.

A typical CCG characterizes with a large number of playable cards (over 1,000 in TESL and almost 20,000 in MtG), which causes an enormous number of possible deck compositions. This leads to even bigger in-play search space, as the player does not know the order of his next draws, the content of the opponent's deck, nor his in-hand cards. Such numbers tend to increase the amount of problems related with the imperfect information and randomness.

It is also common to observe a \emph{metagame} level of such games. It describes the popularity of certain decks or cards. On a top-level, meta creates a possibility to compare different decks on a larger scale. Most often, it boils down to a ``rock-paper-scissors'' scheme, but with more possible types.

\subsection{Related Work}

The problem of creating decks that will be effective for some given meta (usually understood as a currently dominating set of opposing decks) is one of the key challenges for CCG domain \cite{hoover2019many}. 
In \cite{garcia2016evolutionary}, the Hearthstone decks are evolved and tested for their strength via playing against a small number of predefined human-created decks. A similar task, but in a much more complicated domain of Magic: The Gathering has been approached, also via evolution, in \cite{bjorke2017deckbuilding}. A neural network-based approach to deckbuilding in Hearthstone has been presented in \cite{chen2018q}. More in-depth analysis of Hearthstone deck space and the impact of various factors on the process of their evolution can be found in \cite{bhatt2018exploring}. All the above research is focused on the \emph{constructed} game mode, i.e., a static card selection, from unrestricted sets of available cards.

The topic closely correlated with deckbuilding is how to design the cards in CCG to be balanced \cite{volz2016demonstrating}. In \cite{Silva2019evolving}, evolution is used to propose changes to the cards that will result in better balance.
In \cite{Fontaine2019Mapping}, several experiments using a modification of MAP-Elites algorithm for design and rebalancing of Hearthstone have been presented.

So far, no research has been aimed at the arena mode of CCGs.
Methods for estimating card values that could be useful in arena play have not been subject to a proper investigation, although they are popular among human game players. There exist dedicated web pages and game-helping software that recommends cards (e.g., \cite{Lightforge2016,HearthArena2017}). The data they are based on is continuously updated and consists of a mix of expert domain knowledge, mathematical formulas, and remarks made by players on public forums.

\subsection{Legends of Code and Magic}

\emph{Legends of Code and Magic} (LOCM) \cite{LOCMPage} is a small implementation of a Collectible Card Game, designed to perform AI research. Its advantage over the real card game AI engines is that it is much simpler to handle by the agents, and thus allows testing more sophisticated algorithms and quickly implement theoretical ideas.

\begin{figure}
\centering
\includegraphics[width=\linewidth]{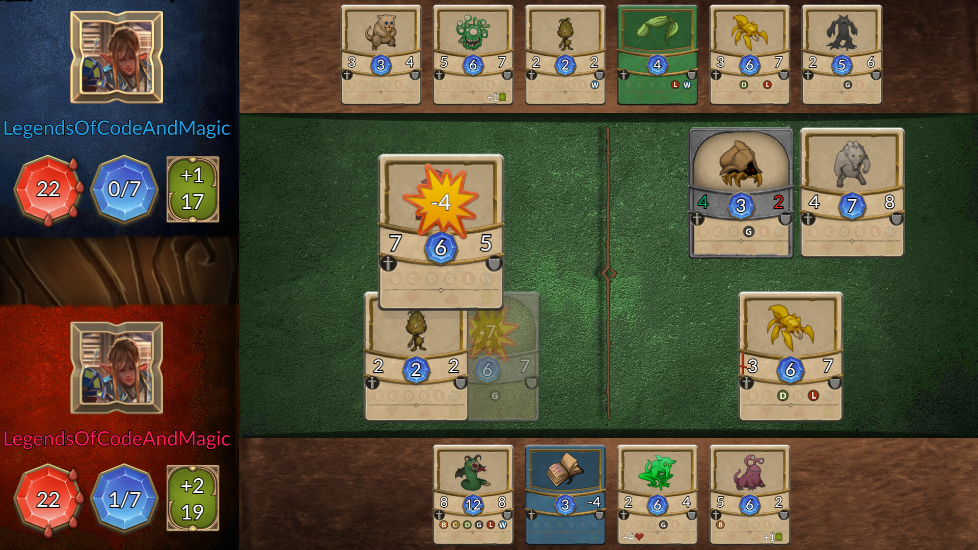}
\caption{Legends of Code and Magic -- in-game visualization.}
\label{fig:locm}
\end{figure}

All card effects are deterministic. Thus the non-determinism is introduced only by the ordering of cards and unknown opponent's deck. The game board consists of two lanes (similarly as in The Elder Scrolls: Legends), so it favors deeper strategic thinking. Fig.~\ref{fig:locm} shows the visualization in the middle of the game. Also, LOCM is based on the fair arena mode, i.e., before every game, both players create their decks secretly from the symmetrical yet limited choices.
The card choices for the players are different every game, but both players have the same decisions in this phase. This is not true in arena mode in existing computer games, where every created deck is used in several games versus players that had other options to choose from.
The deckbuilding in LOCM is more dynamic and, although the concept of meta is still applicable, it can be countered by the specific choices of drafts, reducing the overall strength of usual human-created top-meta decks.

The game in a slightly simplified (one-lane) form was used in August 2018 as a CodinGame platform contest, attracting more than 2,000 players (or rather AI programmers) across the world \cite{LOCMCG}. The \emph{Strategy Card Game AI Competition} based on LOCM was hosted in 2019 by IEEE CEC and IEEE COG \cite{LOCMPage}.

\section{Methodology} 

\subsection{Problem Specification}

Consider two players, $A$ and $B$, playing in a constructed mode. Before every game, they have to choose their decks, i.e., subsets of available cards additionally fulfilling some game-specific constraints (e.g., no more than three copies of the same card). Then, $A$ and $B$ play against each other, using their playing algorithms and their chosen decks (randomly shuffled). 

Thus, the goal of $A$ is to choose deck and algorithm such that it performs best over every possible combination of the opponent's choices. (In practice, as the number of such combinations is huge, a much smaller set of meta opponent decks is considered, as they are well-performing and have a higher chance of appearance.)

However, considering the arena mode, the task becomes slightly different. Now, the player $A$ has to build his deck given a set of choices in the so-called \emph{draft} phase. Usually, it consists of turns in which $A$ has to pick one of randomly given cards, and he is not aware of the future options.
Thus, the player's goal is to choose the best draft strategy, i.e., such that performs best for every draft options and every possible opponent.

In the deckbuilding problem, we fix the algorithms used by the players, treating them as given, and focus on optimizing the policy of choosing the cards. Either a static as in constructed mode, or dynamic, depending on the given draft options, in arena.

\subsubsection{Domain}

Consider a set of all possible cards in the game $\mathbb{C}$. For Legends of Code and Magic, $|\mathbb{C}|=160$. During the draft phase, a player collects 30 cards for his deck in turns, in each turn choosing one of 3 given cards. Thus, given that all draws are independent and a card cannot be repeated within a single turn, the number of possible drafts for LOCM is \\
$(160\times 159\times 158)^{30}\approx 1.33\times 10^{198}$.

During the draft, a player knows his previous decisions but is unaware of future choices. Thus, when learning the best draft policy, we search for the best function from the following domain (simplified to repeating cards):
\begin{equation*}
(\mathbb{C}\cup\{\bot\})^{30}\times\mathbb{C}^3\to\{1,2,3\},
\end{equation*}

where $\bot$ denotes a choice that is yet to make (remaining draft turns). We assume the cards to choose from are ordered, so it is enough to pick the position of the card.

This task can be significantly simplified by discarding the information about previous choices (which removes, e.g., taking to account card synergies or controlling a mana curve), and reduces the domain of possible draft strategies to
\begin{equation*}
 \mathbb{C}^3\to\{1,2,3\}.
\end{equation*}

In this work, for practical reasons, we represent a draft policy as the pure card-value assignment: in each turn, the card with the highest value is chosen. 
Such an approach is popular as a base for the, e.g., Hearthstone arena helpers \cite{Lightforge2016,HearthArena2017}.
It also makes the encoding of the genome easy and relatively small (vector of $|\mathbb{C}|$ real numbers), while still providing a relatively good estimation of real-card value given that learning is performed with respect to fixed playing algorithms.

\subsubsection{Fitness Function}

The difficulty with estimating the quality of card selection policy comes from the multiple sources of simplifying assumptions and randomness. 
Even approximated fitness functions, using fixed playing algorithms, are flawed not only because of unknown distribution of opponent strategies, but also the nondeterminism in the games themselves, causing the value of the function to be less reliable.

In such a case, a reasonable approach is to combine simulation-based estimations with numbers. First, sample the largest possible portion of possible drafts. 
Second, estimate fitness based on the performance of candidate policies on those sample drafts, using a large number of simulated games. 

\begin{figure}
\centering
\def\svgwidth{\columnwidth}
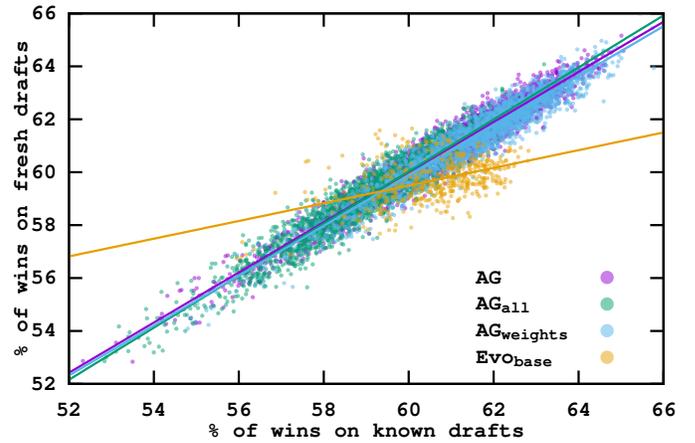
\caption{Correlation between performance during training (x axis) and evaluation (y axis). All algorithms were trained on 100 drafts and evaluated on 250 drafts. A single evaluation consisted of 250 games against two baselines and three random strategies using the random player. Clearly, $\mathit{Evo_{base}}$ generalizes worse than any of $\mathit{AG}$ algorithms, which performance stays at the same level.}
\label{fig:correl}
\end{figure}

To be sound, this approach requires that the performance on a low number of training samples correlate with the performance on a larger set of unseen test samples.
To ensure this is true, we performed additional tests on draft strategies of various quality and obtained from various sources. 
The results, presented in Fig.~\ref{fig:correl}, confirms such a correlation.


\subsection{Simple Baselines and Measuring Computational Cost}

To compare the algorithms fairly, we introduce a cost measure $\mathcal{C}$, which is equal to the number of simulated games required to compute the solution.
This is more accurate than simple generation-based comparison, as the total cost of calculating one generation may differ between two algorithms.

We introduce the following notation: $n$ is the size of a population, $d_t$ is the number of random drafts used to train the solution, $s_g$ is the number of games played when comparing two individuals.

To provide a simple baseline for the quality of evolution, we use two types of randomly generated solutions. 
First, called $\mathit{Random_{all}}$, creates $n$ random individuals and plays $s_g$ games (with side change) on every training draft between every pair of the individuals.
An individual which won the most games is the solution. The computation cost of this method is:
\begin{equation*}
\mathcal{C}_{\mathit{Random_{all}}} = n \times (n-1) \times s_g \times d_t.
\end{equation*}

The second random baseline, called $\mathit{Random_{tournament}}$, randomizes $n$ individuals and performs a tournament to select the best representative. To determine the winner of each matchup, $s_g$ games on every training draft are played. Thus the computation cost is:
\begin{equation*}
\mathcal{C}_{\mathit{Random_{tournament}}} = (n-1) \times s_g \times d_t.
\end{equation*}

We also use two precomputed card orderings, established by some of the top participants of the one-lane CodinGame LOCM contest \cite{CG2018LOCMPostmortem}. Thus, they have no computation cost. We call this strategies $\mathit{\mathit{Contest_1}}$ and $\mathit{\mathit{Contest_2}}$.

\subsection{Baseline Evolutionary Algorithm}

The goal of the evolution is to improve the quality of draft policies encoded as chromosomes. A straightforward approach is to use all available training data in each generation. 

A genotype is represented as a constant-size vector of doubles. Every gene (from 1 to $|\mathbb{C}|=160$) encodes a priority of the associated card (estimated card value). Thus, during the draft phase, a card with the highest priority is chosen each~turn. 

As before, let $n$ be the size of the population, and $d_t$ the number of available training drafts. The schema of the baseline evolutionary algorithm $\mathit{Evo_{base}}$ goes as follows.

The population is initialized with the random values between 0.0 and 1.0.
To evaluate the population, we play $s_g$ games between every two individuals for each of $d_t$ drafts (with side change after half of the games).
Then, we use tournament selection (best of random 4 individuals) to select $\frac{n}{2}$ parents based on the number of wins.
The standard uniform crossover is used. Mutation rate $m=0.05$ is the probability of changing each gene into a new random number. The elitism size is 2.
Thus, to compute $g$ generations, this algorithm has to play $\mathcal{C}_{\mathit{Evo_{base}}}$ games, where
\begin{equation*}
\mathcal{C}_{\mathit{Evo_{base}}} = n \times (n-1) \times s_g \times d_t \times (1 + g).
\label{eq:CEb}
\end{equation*}

\section{Active Genes Algorithm}

Evaluating each generation using all of the available test drafts, while being the most robust, is also time-consuming and may force the evolution to be insubstantial within the constrained computational budget.

Alternatively, we propose an approach where each generation is responsible for learning how to play only one of the available test drafts.
Such a method allows not only evolving more generations within the same budget but also observing a more detailed influence of single genes (cards) on the performance of the agents in particular scenarios (drafts). 
To emphasize the benefit of this gene-to-draft-to-outcome correspondence even more, and partially make up for the loss of generality it creates, evolutionary operators will be applied selectively, differently in each generation.

We say a gene is \emph{active} in a given generation if the card it encodes appears within the draft choices.
For the sake of evaluation of this generation, all the other genes are considered irrelevant.
Thus, in this generation, we can perform crossover and mutation only on the active genes.
Moreover, this will prevent the destruction of the so-far gained knowledge encoded in the inactive genes, which has already proven itself through the previous generations.

The major drawback of this approach is that we lose a uniform metric that can be used to compare parents and children across the generations.
Thus, to select $\frac{n}{2}$ pairs of parents, we run actual games to test how they perform on the current generation draft.
For each parent we perform a tournament of size $s_{tSize}=4$, playing $s_{tGames}=10$ games each round.
Selected parents create offspring in the same manner as in the $\mathit{Evo_{base}}$ algorithm (uniform crossover, constant mutation rate).

Now, we calculate fitness values for the offspring population.
This is done in $s_r$ rounds, where in each round we score the population by the so-far wins and then play $s_g$ games (with side change) between consecutive pairs in order.


The remaining part, selection, is in our approach substituted by a \emph{merge} operation.
To create each of $n$ individuals of the next population, we use a fitness-based roulette to select a child from the offspring, and a parent from the parents' population.
For parents, the fitness values of their generation are used.

We investigated three variants of the merging procedure.
First, called $\mathit{AG}$, uses active genes most straightforwardly.
The resulting individual contains values of the active genes copied from the child and inherits the rest from its parent.
Thus, the newest knowledge is considered the most important.

Alternatively, $\mathit{AG_{weights}}$ is the variant of $\mathit{AG}$ that uses the weighted sum instead.
Such a variant aims to be more conservative and improve individuals gradually, preserving the already gained knowledge even more.
The proportion we found working is $0.75$ of the parent gene and $0.25$ of the child gene, instead of the original $0$ and $1$ in $\mathit{AG}$.

For comparison, we also test $\mathit{AG_{all}}$ variant, which is based on the same scheme but does not take advantage of active genes (i.e., it treats all genes as active).
Instead, it discards the parent, and only the child gene values are copied.

Essentials of the pseudocode for these algorithms are presented in Fig.~\ref{fig:pseudocode}.
Here, we assume that $g=d_t$, which is in align with most of the conducted experiments.
More sophisticated variants, using the same pseudocode but breaking this assumption, are described separately in Section~\ref{sec:agapproaches}.
The main procedure, \textsc{Evolve}, progresses through each generation learning $d_t$ drafts, creating random drafts using \textsc{RandomDrafts}, calculating the children population using \textsc{CreateOffspring}, and merging them, as described above, using \textsc{MergeAll}.
Depending on the variant, \textsc{MergeOne} selects an appropriate behavior, where \textsc{lerp} is the linear interpolation function.
The \textsc{Score} procedure simulates a single game and updates the scores (win counts) accordingly.

Thus, the cost of $g$ generations using these algorithms is:
\begin{equation*}
\begin{gathered}
\mathcal{C}_{\mathit{AG}} = \mathcal{C}_{\mathit{AG_{weights}}} = \mathcal{C}_{\mathit{AG_{all}}} = \\
n \times g \times (s_{tSize} \times (s_{tSize} - 1) \times s_{tGames} + s_r \times s_g \times d_t)
\label{eq:CEaw}
\end{gathered}
\end{equation*}


\begin{algorithm}[]
\caption{Active genes algorithm variants pseudocode.}
\label{fig:pseudocode}
\begin{algorithmic}
  \Procedure{Evolve}{options}
    \State{old $\gets$ \Call{RandomPopulation}{$n$}}
    \For{generation $\gets$ 1, $g$}
      \State{drafts $\gets$ \Call{RandomDrafts}{$ $}}
      \State{new $\gets$ \Call{CreateOffspring}{drafts, old}}
      \State{old $\gets$ \Call{MergeAll}{old, new, drafts, options}}
    \EndFor
    \State{\Return old}
  \EndProcedure
  \\
  \Procedure{CreateOffspring}{drafts, old}
    \State{new $\gets$ \Call{EmptyPopulation}{$ $}}
    \For{individual $\in$ new}
      \State{parents $\gets$ \Call{SelectParents}{drafts, old}}
      \State{children $\gets$ \Call{Crossover}{parents}}
      \State{individual $\gets$ \Call{Mutate}{children}}
    \EndFor
    \State{\Call{ScorePopulation}{drafts, new}}
    \State{\Return new}
  \EndProcedure
  \\
  \Procedure{SelectParents}{drafts, population}
    \State{tournament $\gets$ \Call{Sample}{population, $s_{tSize}$}}
    \ForAll{draft $\in$ drafts}
      \ForAll{a $\in$ tournament}
        \ForAll{b $\in$ tournament}
          \If{a $\neq$ b}
            \For{game $\gets$ 1, $s_{tGames}$ / 2}
              \State{\Call{Score}{a, b, draft}}
              \State{\Call{Score}{b, a, draft}}
            \EndFor
          \EndIf
        \EndFor
      \EndFor
    \EndFor
    \State{\Return \Call{Best2ByScore}{tournament}}
  \EndProcedure
  \\
  \Procedure{ScorePopulation}{drafts, population}
    \For{round $\gets$ 1, $s_r$}
      \State{\Call{SortByScore}{population}}
      \ForAll{draft $\in$ drafts}
        \For{i $\gets$ 1, 3, 5, \dots,  $n$}
          \State{a, b $\gets$ population[i], population[i + 1]}
          \State{\Call{Score}{a, b, draft}}
          \State{\Call{Score}{b, a, draft}}
        \EndFor
      \EndFor
    \EndFor
  \EndProcedure
  \\
  \Procedure{MergeAll}{old, new, drafts, options}
    \State{merged $\gets$ \Call{EmptyPopulation}{$ $}}
    \ForAll{individual $\in$ merged}
      \State{a, b $\gets$ \Call{Roulette}{old}, \Call{Roulette}{new}}
      \State{individual $\gets$ \Call{MergeOne}{a, b, drafts, options}}
    \EndFor
    \State{\Return merged}
  \EndProcedure
  \\
  \Procedure{MergeOne}{new, old, drafts, options}
    \If{$\mathit{AG_{all}}$}
      \State{cardIds $\gets$ \textproc{AllCardIds}}
    \Else
      \State{cardIds $\gets$ \Call{CardIdsIn}{drafts}}
    \EndIf
    \State{merged $\gets$ \Call{Clone}{old}}
    \ForAll{id $\in$ cardIds}
      \State{old[id] $\gets$ \Call{lerp}{new[id], old[id], options.weight}}
    \EndFor
    \State{\Return merged}
  \EndProcedure
\end{algorithmic}
\end{algorithm}

\section{Experiments}

Source code for the described experiments is available in a public GitHub repository \cite{ActiveGenesRepo}. All performed experiments were run on a single CPU-Optimized DigitalOcean droplet with 16 GB of RAM and 8 standardized vCPUs.




\subsection{Algorithm Comparison}

We compared the overall results obtained by the $\mathit{\mathit{Contest_1}}$, $\mathit{\mathit{Contest_2}}$, $\mathit{Random}$, $\mathit{Random_t}$, $\mathit{Evo_{base}}$, $\mathit{AG}$, $\mathit{AG_{all}}$, and $\mathit{AG_{weights}}$ algorithms.
To ensure the robustness of our approach, we have tested two playing strategies -- random and greedy.
We did not test more advanced strategies due to the computational time constraints.

The first one uniformly picks a random action sequence, while the latter selects the best actions, one at a time, according to a material heuristic.
A complete summary of their relative performance is presented in Tables \ref{tab:algcomp-random} and \ref{tab:algcomp-greedy}.

While the competition-based heuristic $\mathit{\mathit{Contest_1}}$ gains fewer wins than loses, $\mathit{\mathit{Contest_2}}$ seems to be the second most robust choice from considered strategies.
What is interesting, during the one-lane LOCM contest \cite{CG2018LOCMPostmortem} $\mathit{\mathit{Contest_1}}$ was established as the meta, outperforming other solutions, including $\mathit{\mathit{Contest_2}}$.

Scores of both random players are worse than those obtained by any evolution-based approach, which is a clear indication that the learning process gives a definite advantage.

\begin{figure*}
\centering
\def\svgwidth{\columnwidth}
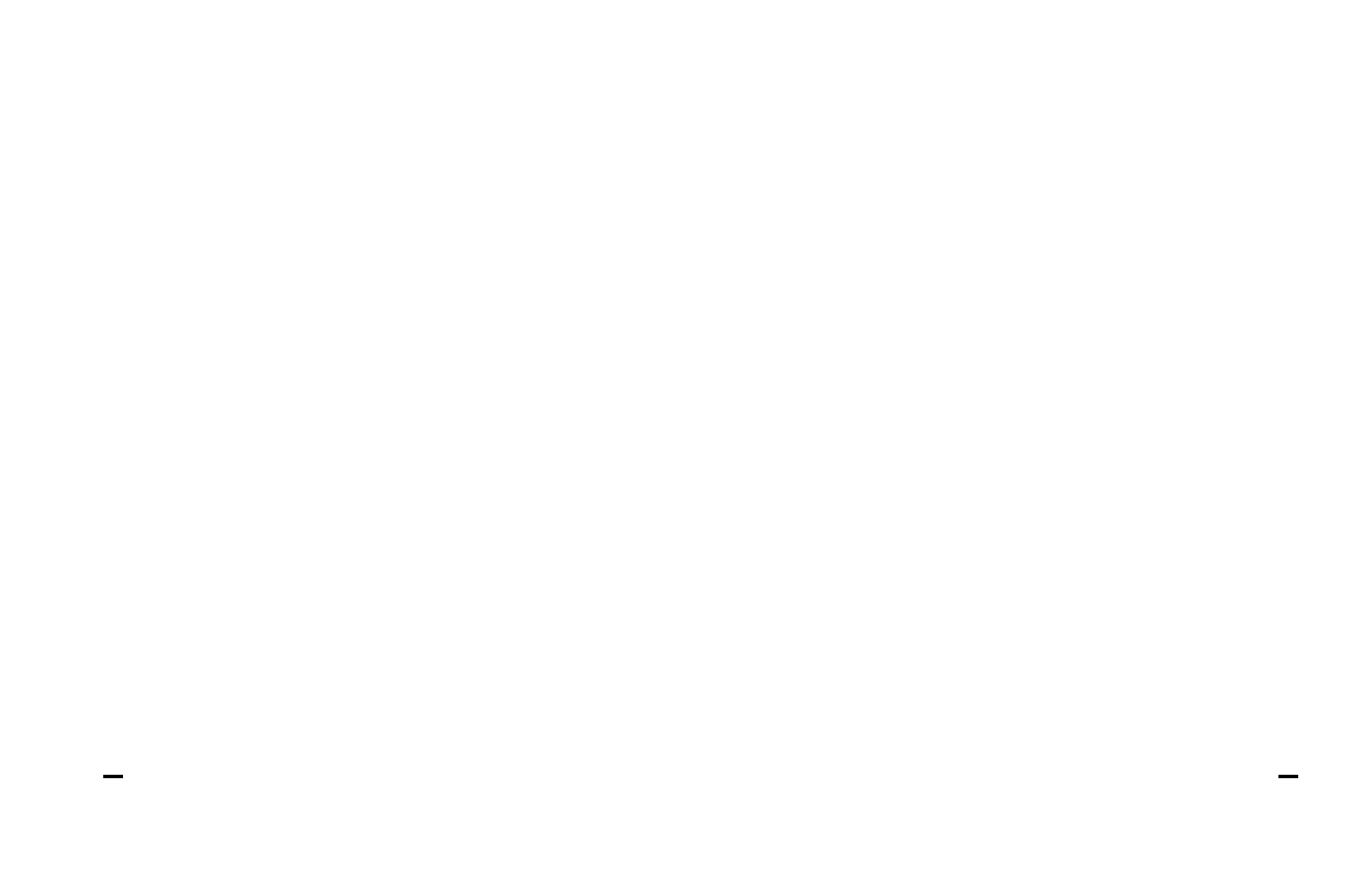
\def\svgwidth{\columnwidth}
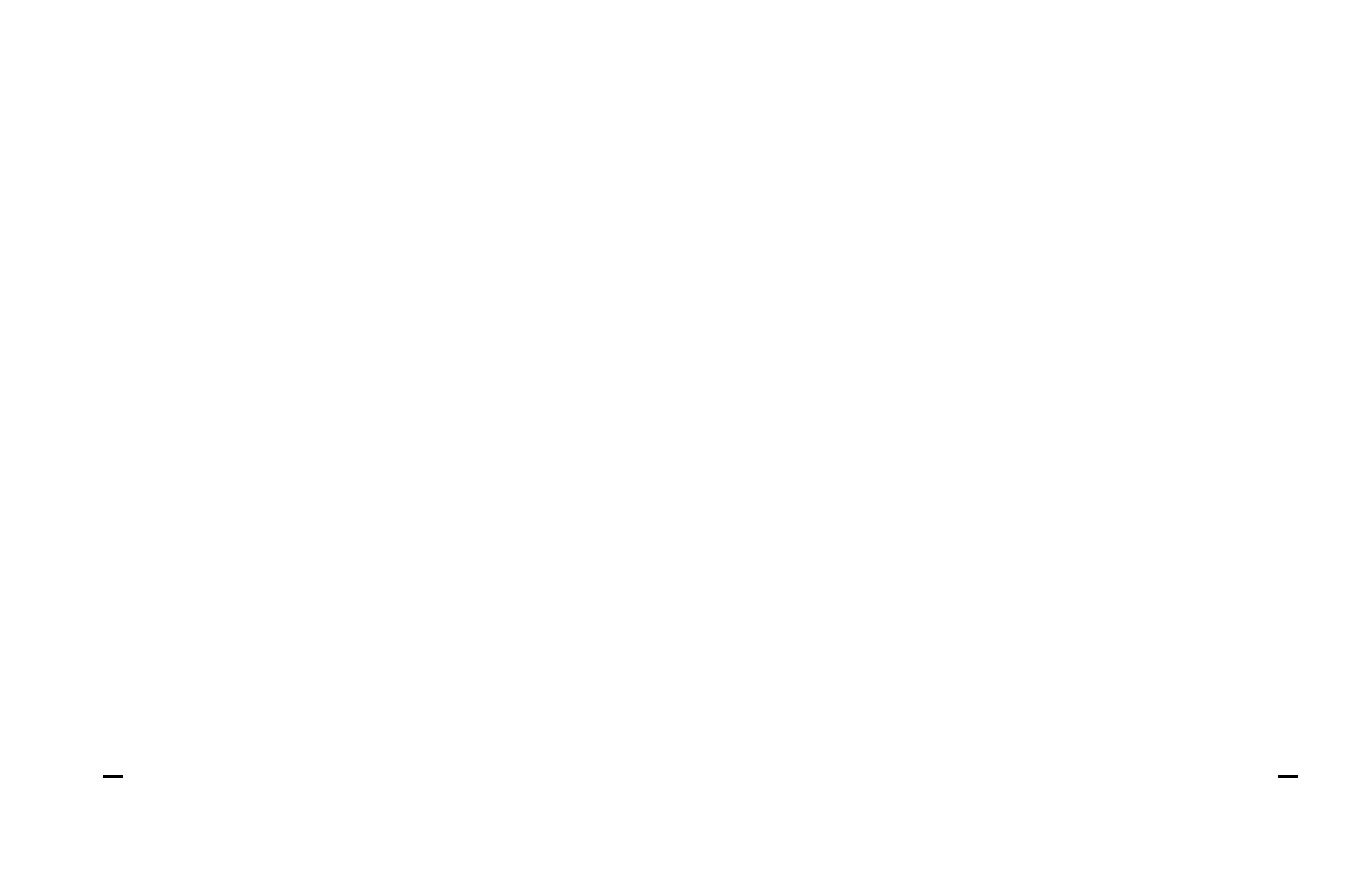
\caption{Process of evolution for $\mathit{AG}$, $\mathit{AG_{all}}$, $\mathit{AG_{weights}}$, and $\mathit{Evo_{base}}$ using random (left chart) and greedy (right chart) players. Computation cost (number of simulated games) on x axis, average performance versus $\mathit{\mathit{Contest_1}}$, $\mathit{\mathit{Contest_2}}$, and three random strategies on y axis.}
\label{fig:algcomp}
\end{figure*}

Additionally, on Fig.~\ref{fig:algcomp}, we visualized the process of evolution for $\mathit{Evo_{base}}$, $\mathit{AG}$, $\mathit{AG_{all}}$, and $\mathit{AG_{weights}}$.
For each of the presented algorithms, the best five individuals from each generation played $50$ games on $250$ random drafts against $\mathit{\mathit{Contest_1}}$, $\mathit{\mathit{Contest_2}}$, and three random individuals.
The results reported on this chart are higher than in the tables mentioned before, as the random opponents tend to be less skilled on average.

Surprisingly, the performance of the straightforward evolution $\mathit{Evo_{base}}$ is below $50\%$ when using the random player. Also, the correlation between performance on the training and evaluation drafts is the worst (yellow line on the Fig.~\ref{fig:correl}).
Each of its generations took significantly longer to compute, so only a few of them can be finished within the assumed computation budget.
It might be the case that their number is too low to observe learning.
Nevertheless, a few non-exhaustive experiments we had additionally performed showed that the score does not raise significantly even with a far greater computational budget or using different parameters.
This variant generates average solutions during the first generation that it cannot further improve.

As expected, $\mathit{AG_{all}}$ performs poorly. Using only offspring genes results in constant forgetting, which is visible in its evolution process.
On the contrary, the remaining active genes-based approaches, $\mathit{AG}$ and $\mathit{AG_{weights}}$, learn step-by-step from low scores.
Both need about four times the cost of the initial $\mathit{Evo_{base}}$ generation to start performing better, but the process does not finish there.
They continue learning, which supports the assumption of the advantage of batched learning and selective genetic operators.
The variant with weighted merge achieves significantly better results versus all opponents, especially using the greedy playing algorithm.
In particular, it tends to achieve better performance faster, as it is easier to stabilize at good gene values by weighted sum than it is by gene replacing.

\begin{table*}
\caption{A comprehensive comparison of all algorithms using random player. Each was trained $10$ times with a computational budget of 1,000,000, yielding $50$ best players. Each two played $20$ games on $500$ random drafts. The whole experiment was repeated $5$ times. All scores are averaged, followed by their standard variations. The best results of each column are in bold.}
\label{tab:algcomp-random}
\setlength{\tabcolsep}{4pt}
\begin{tabular}{ l c c c c c c c c c c c }
                        &      $\mathit{\mathit{Contest_1}}$ &      $\mathit{\mathit{Contest_2}}$ &       $\mathit{AG_{all}}$ &   $\mathit{AG_{weights}}$ &             $\mathit{AG}$ &     $\mathit{Evo_{base}}$ &         $\mathit{Random}$ &       $\mathit{Random_t}$ &                   Average \\
\hline
$\mathit{\mathit{Contest_1}}$    &                       $-$ &          48.73 $\pm$ 0.60 &          50.54 $\pm$ 0.25 &          48.35 $\pm$ 0.32 &          49.30 $\pm$ 0.29 &          50.43 $\pm$ 0.12 &          51.38 $\pm$ 0.19 &          51.23 $\pm$ 0.15 &          49.88 $\pm$ 0.14 \\
$\mathit{\mathit{Contest_2}}$    &          51.27 $\pm$ 0.60 &                       $-$ &          51.95 $\pm$ 0.25 & \textbf{49.67} $\pm$ 0.24 &          50.70 $\pm$ 0.28 &          52.12 $\pm$ 0.20 &          52.87 $\pm$ 0.20 &          52.75 $\pm$ 0.20 &          51.46 $\pm$ 0.09 \\
$\mathit{AG_{all}}$     &          49.46 $\pm$ 0.25 &          48.04 $\pm$ 0.25 &                       $-$ &          47.70 $\pm$ 0.09 &          48.49 $\pm$ 0.06 &          49.85 $\pm$ 0.08 &          50.78 $\pm$ 0.05 &          50.59 $\pm$ 0.05 &          49.16 $\pm$ 0.06 \\
$\mathit{AG_{weights}}$ & \textbf{51.64} $\pm$ 0.32 & \textbf{50.32} $\pm$ 0.24 & \textbf{52.29} $\pm$ 0.09 &                       $-$ & \textbf{50.93} $\pm$ 0.07 & \textbf{52.27} $\pm$ 0.03 & \textbf{53.08} $\pm$ 0.05 & \textbf{52.99} $\pm$ 0.04 & \textbf{51.74} $\pm$ 0.04 \\
$\mathit{AG}$           &          50.69 $\pm$ 0.29 &          49.29 $\pm$ 0.28 &          51.50 $\pm$ 0.06 &          49.06 $\pm$ 0.07 &                       $-$ &          51.43 $\pm$ 0.07 &          52.33 $\pm$ 0.03 &          52.11 $\pm$ 0.04 &          50.76 $\pm$ 0.05 \\
$\mathit{Evo_{base}}$   &          49.56 $\pm$ 0.12 &          47.87 $\pm$ 0.20 &          50.14 $\pm$ 0.08 &          47.72 $\pm$ 0.03 &          48.56 $\pm$ 0.06 &                       $-$ &          50.93 $\pm$ 0.04 &          50.69 $\pm$ 0.03 &          49.24 $\pm$ 0.04 \\
$\mathit{Random}$       &          48.61 $\pm$ 0.19 &          47.12 $\pm$ 0.20 &          49.22 $\pm$ 0.05 &          46.91 $\pm$ 0.05 &          47.66 $\pm$ 0.03 &          49.06 $\pm$ 0.05 &                       $-$ &          49.82 $\pm$ 0.03 &          48.26 $\pm$ 0.02 \\
$\mathit{Random_t}$     &          48.76 $\pm$ 0.14 &          47.24 $\pm$ 0.20 &          49.40 $\pm$ 0.05 &          47.00 $\pm$ 0.04 &          47.88 $\pm$ 0.04 &          49.30 $\pm$ 0.03 &          50.17 $\pm$ 0.03 &                       $-$ &          48.44 $\pm$ 0.02
\end{tabular}
\end{table*}

\begin{table*}
\caption{A comprehensive comparison of all algorithms using greedy player. Each was trained $3$ times with a computational budget of 1,000,000, yielding $50$ best players. Each two played $20$ games on $500$ random drafts. The whole experiment was repeated $3$ times. All scores are averaged, followed by their standard variations. The best results of each column are in bold.}
\label{tab:algcomp-greedy}
\setlength{\tabcolsep}{4pt}
\begin{tabular}{ l c c c c c c c c c c c }
                        &      $\mathit{\mathit{Contest_1}}$ &      $\mathit{\mathit{Contest_2}}$ &       $\mathit{AG_{all}}$ &   $\mathit{AG_{weights}}$ &             $\mathit{AG}$ &     $\mathit{Evo_{base}}$ &         $\mathit{Random}$ &       $\mathit{Random_t}$ &                   Average \\
\hline
$\mathit{\mathit{Contest_1}}$    &                       $-$ &          51.48 $\pm$ 0.44 &          48.27 $\pm$ 0.20 &          40.75 $\pm$ 0.25 &          42.72 $\pm$ 0.35 &          44.73 $\pm$ 0.44 &          50.49 $\pm$ 0.17 &          51.89 $\pm$ 0.67 &          47.19 $\pm$ 0.17 \\
$\mathit{\mathit{Contest_2}}$    &          48.52 $\pm$ 0.44 &                       $-$ &          48.81 $\pm$ 0.56 &          42.05 $\pm$ 0.39 &          44.21 $\pm$ 0.37 &          45.94 $\pm$ 0.39 &          52.30 $\pm$ 0.16 &          53.61 $\pm$ 0.38 &          47.92 $\pm$ 0.11 \\
$\mathit{AG_{all}}$     &          51.73 $\pm$ 0.20 &          51.19 $\pm$ 0.56 &                       $-$ &          43.38 $\pm$ 0.25 &          45.34 $\pm$ 0.03 &          48.14 $\pm$ 0.33 &          52.77 $\pm$ 0.14 &          54.82 $\pm$ 0.17 &          49.62 $\pm$ 0.11 \\
$\mathit{AG_{weights}}$ & \textbf{59.25} $\pm$ 0.25 & \textbf{57.95} $\pm$ 0.39 & \textbf{56.62} $\pm$ 0.25 &                       $-$ & \textbf{51.95} $\pm$ 0.12 & \textbf{54.75} $\pm$ 0.25 & \textbf{59.33} $\pm$ 0.06 & \textbf{61.09} $\pm$ 0.18 & \textbf{57.28} $\pm$ 0.06 \\
$\mathit{AG}$           &          57.28 $\pm$ 0.35 &          55.79 $\pm$ 0.37 &          54.66 $\pm$ 0.03 & \textbf{48.05} $\pm$ 0.12 &                       $-$ &          52.91 $\pm$ 0.12 &          57.84 $\pm$ 0.15 &          59.58 $\pm$ 0.16 &          55.16 $\pm$ 0.12 \\
$\mathit{Evo_{base}}$   &          55.27 $\pm$ 0.44 &          54.06 $\pm$ 0.39 &          51.86 $\pm$ 0.33 &          45.25 $\pm$ 0.25 &          47.09 $\pm$ 0.12 &                       $-$ &          54.96 $\pm$ 0.03 &          56.67 $\pm$ 0.09 &          52.16 $\pm$ 0.19 \\
$\mathit{Random}$       &          49.51 $\pm$ 0.17 &          47.70 $\pm$ 0.16 &          47.23 $\pm$ 0.14 &          40.67 $\pm$ 0.06 &          42.16 $\pm$ 0.15 &          45.04 $\pm$ 0.03 &                       $-$ &          51.97 $\pm$ 0.10 &          46.32 $\pm$ 0.02 \\
$\mathit{Random_t}$     &          48.11 $\pm$ 0.67 &          46.39 $\pm$ 0.38 &          45.18 $\pm$ 0.17 &          38.91 $\pm$ 0.18 &          40.42 $\pm$ 0.16 &          43.33 $\pm$ 0.09 &          48.03 $\pm$ 0.10 &                       $-$ &          44.34 $\pm$ 0.15
\end{tabular}
\end{table*}

\begin{figure*}
\centering
\def\svgwidth{0.9\textwidth}
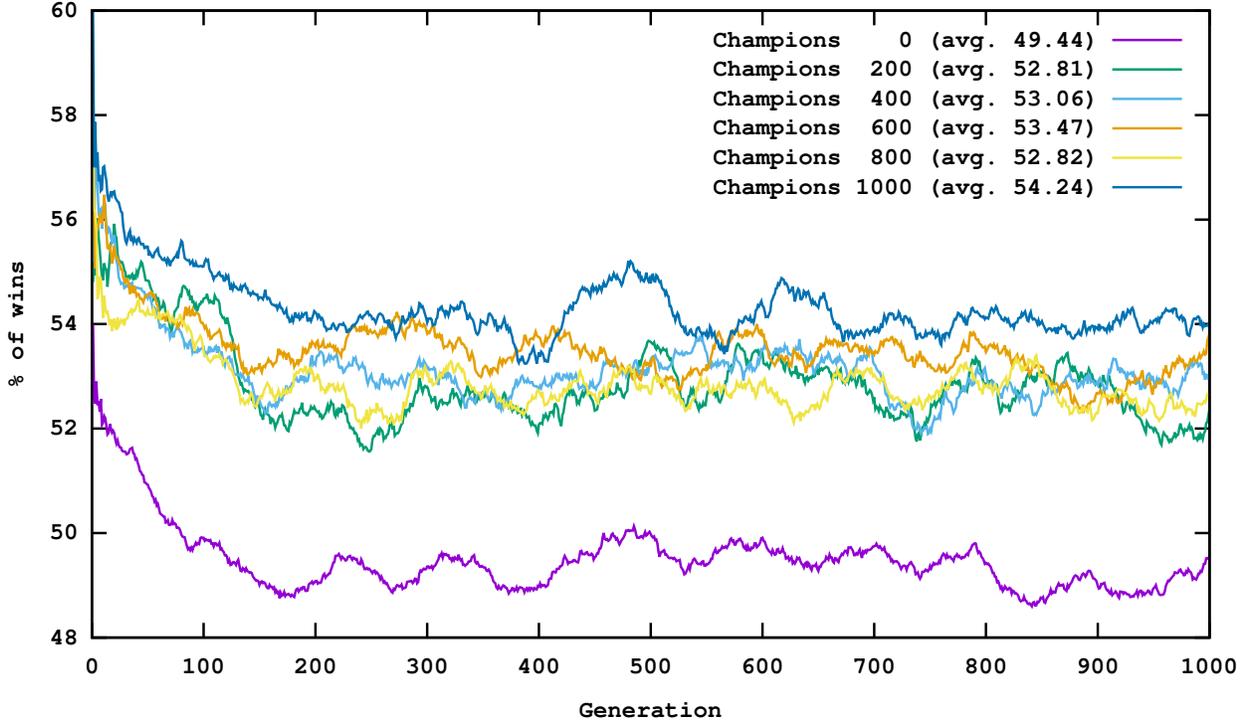
\caption{Average performance of $\mathit{AG_{weights}}$ champions against best five individuals of each generation. Each group played $50$ games on every of the $1000$ training drafts playing randomly. The average win rate (y axis) tends to drop, as the champions of the following generations (x axis) are getting stronger.}
\label{fig:specchamps}
\end{figure*}

\begin{figure}
\centering
\def\svgwidth{\columnwidth}
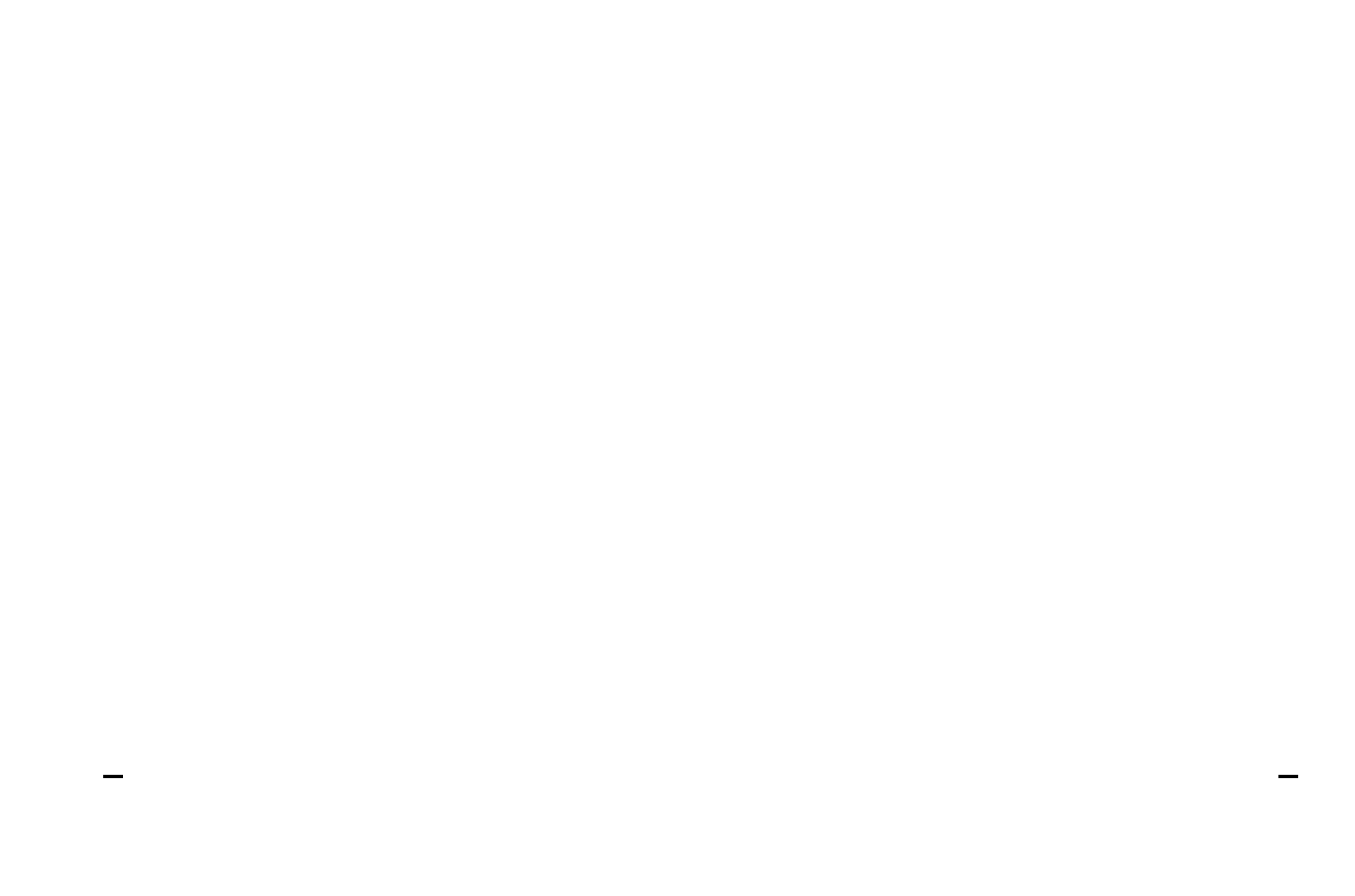
\caption{Process of evolution for $\mathit{AG_{weights}}$ variants using random player. Computation cost (number of simulated games) on x axis, average performance versus $\mathit{\mathit{Contest_1}}$, $\mathit{\mathit{Contest_2}}$, and three random strategies on y axis.}
\label{fig:algcomp-variants}
\end{figure}

Although the improvements of a single percent using random players do seem small, it is a common thing in such a noisy environment as CCGs, and even that leads to a long-term gain.
Especially, when it translates into more significant improvement for better player strategies.
For comparison, using greedy players for both evolution and evaluation leads to more diversified results, emphasizing the difference between the learning algorithms.

\newpage

Moreover, a comparison between Tab.~\ref{tab:algcomp-random}, Tab.~\ref{tab:algcomp-greedy} and Fig.~\ref{fig:algcomp} shows that even a small advantage over the better performing opponents yields a significant advantage over the less skilled ones.

\subsection{Analysis of $\mathit{AG_{weights}}$ Learning}

In Fig.~\ref{fig:specchamps}, we visualized the performance of overall-best against generation-best individuals during a sample $\mathit{AG_{weights}}$ run to analyze the learning progress.
It shows how the all-time top five individuals per $200$ generations (the first generation is an entirely random one) perform against the top five individuals of each of the generations on all training drafts.

As we can observe, champions from later generations tend to perform better on average, which is consistent with the previous observations.
All tested champions are significantly strong at first because the first generation did not have time to learn.
As the learning progresses and the individuals from the following generations are getting better, the scores of the champions tend to be lowering.

It is worth noticing that the best genotypes of each generation were chosen based on the current generation draft, but the chart shows their performance on all $d_t$ drafts.
Thus, when we treat champions as constant opponents that differ in strength, we can observe changes in the overall score after learning each new training draft.
Each descent means that it improves overall performance, while each rise signals that the overall performance decreased, even though only one particular draft was examined.
\newpage 
In many places, both peaks and valleys are in similar places for multiple champion lines, which shows the importance of particular drafts.

\subsection{Investigating Active Genes Approach}
\label{sec:agapproaches}

The performance of active genes evolution correlates with the number of training drafts used to evaluate a single population.
More drafts imply that a more significant part of the genotype is active and being affected by the operators.
On the other hand, with a limited number of drafts per generation, the quality of the evaluation is limited, and some knowledge may be lost during evolution.
Thus, we introduced two additional variants of $\mathit{AG_{weights}}$ algorithm.

The first one, called $\mathit{AG_{weights/Kd}}$, where $K$ is the number of drafts, affects parent tournament and merge phases, in both places adding a loop over the used drafts.
To keep the computational cost and the number of drafts the same, this variant runs for $g / K$ generations compared to $\mathit{AG_{weights}}$.

The second variant called $\mathit{AG_{weights/Kg}}$, where $K$ is the number of repetitions, bases on reusing the same drafts $K$ times, leaving the evolution framework as-is.
To ensure comparable computation cost, as the variant runs for $g \times K$ generations, the number of games in each generation is accordingly lower.

Analyzing the results from Fig.~\ref{fig:algcomp-variants}, the average performance of $\mathit{AG_{weights/4g}}$ is the lowest.
The budget allowed too few evaluations to make one-generation learning reliable enough.
However, less restricted variant $\mathit{AG_{weights/2g}}$, although not the top one, performs reasonably well, achieving higher performance than our $\mathit{Evo_{base}}$ baseline evolution without problems (compare with Fig.~\ref{fig:algcomp} for random player).

There is no significant difference between the performance of $\mathit{AG_{weights/2d}}$ and $\mathit{AG_{weights/4d}}$.
Both performs similarly, slightly worse than the two leading algorithms.
When we compare the difference between those approaches in terms of percent of the genome that is active, we get that it is $\sim$56\% for $\mathit{AG_{weights}}$, $\sim$79\% for $\mathit{AG_{weights/2d}}$, and $\sim$95\% for $\mathit{AG_{weights/4d}}$.

Evolution based on active genes usually performs better when the number of such genes is lower, but this trend is less visible in $\mathit{AG_{weights}}$ variant.
In LOCM, proportion of active genes depends on the generation method used to prepare drafts.
(This is similar to the \emph{dropout} regularization technique in the artificial neural networks.)

Additionally, we performed two more experiments concerning the trade-off between the number of generations and the number of plays during the evaluation, using larger $K$ values.
Both yielded similar but noisier results, therefore have been not included in the paper.


\section{Conclusion}
This paper presents initial research towards the problem of deckbuilding in the arena mode of Collectible Card Games. This can be seen as a complementary problem for the standard CCG deckbuilding, where the set of available cards is known in advance.
As the domain is characterized by vast state space and omnipresent non-determinism, a straightforward approach to learn draft strategies via an evolutionary algorithm is not very successful and lefts much room for improvements.

In our work, we propose an \emph{active genes approach}, a variant that learns gradually, generation-to-generation. Learning in each generation is based only on the partial training data, and genetic operators are applied selectively only on a subset of genes that is currently considered as relevant.
The selection operator is substituted by \emph{merge}, performed between selected pairs consisting of parent and offspring.

We have tested our approach in programming game Legends of Code and Magic, which is used in the Strategy Card Game AI Competition. 
We designed a few variants of the algorithm and conducted experiments show that usually they perform better than the baseline. Some of them achieve average results that are even significantly better, taking into account that in such a noisy environment as CCG (especially in arena mode), even a small increase in win percentage is a substantial gain leading to overall success in a longer timeline.

What is also important, most of the presented approaches learn very fast in terms of our cost measure (which is the number of required game simulations). 
Given a fast simulation engine available on the Strategy Card Game AI Competition package \cite{LOCMSource}, it requires about half a minute for a random player to achieve a decent performance on an average run. 

For future work, we mainly plan to test active genes family of algorithms on other similar domains, e.g., in Hearthstone. 
From the game balancing point of view, it would also be interesting to compare policies obtained by one-lane and two-lane versions of LOCM, and how policies trained for one playing algorithm works for the others.

We also aim to develop methods that will improve reliability in terms of achieved performance. 
This includes using more sophisticated simulated agents (at the expense of computation time) and merging a few runs to perform multiple levels of evolution automatically.
We would also like to investigate more algorithm variants and understand what influence on the behavior of the agents has an algorithm that prepares draft choices.
Finally, we would like to test how our approach affects agent performance compared to the other solutions presented at the Strategy Card Game AI Competition.




\bibliographystyle{IEEEtran} 
\bibliography{bibliography} 

\end{document}

%% file: plot-correlation.pdf_tex
\begingroup%
  \makeatletter%
  \providecommand\color[2][]{%
    \errmessage{(Inkscape) Color is used for the text in Inkscape, but the package 'color.sty' is not loaded}%
    \renewcommand\color[2][]{}%
  }%
  \providecommand\transparent[1]{%
    \errmessage{(Inkscape) Transparency is used (non-zero) for the text in Inkscape, but the package 'transparent.sty' is not loaded}%
    \renewcommand\transparent[1]{}%
  }%
  \providecommand\rotatebox[2]{#2}%
  \newcommand*\fsize{\dimexpr\f@size pt\relax}%
  \newcommand*\lineheight[1]{\fontsize{\fsize}{#1\fsize}\selectfont}%
  \ifx\svgwidth\undefined%
    \setlength{\unitlength}{450bp}%
    \ifx\svgscale\undefined%
      \relax%
    \else%
      \setlength{\unitlength}{\unitlength * \real{\svgscale}}%
    \fi%
  \else%
    \setlength{\unitlength}{\svgwidth}%
  \fi%
  \global\let\svgwidth\undefined%
  \global\let\svgscale\undefined%
  \makeatother%
  \begin{picture}(1,0.66666667)%
    \lineheight{1}%
    \setlength\tabcolsep{0pt}%
    \put(0,0){\includegraphics[width=\unitlength,page=1]{plot-correlation.pdf}}%
    \put(0.07,0.08133333){\color[rgb]{0,0,0}\makebox(0,0)[rt]{\lineheight{1.25}\smash{\begin{tabular}[t]{r}\texttt{\textbf{\footnotesize{52}}}\end{tabular}}}}%
    \put(0,0){\includegraphics[width=\unitlength,page=2]{plot-correlation.pdf}}%
    \put(0.07,0.16083333){\color[rgb]{0,0,0}\makebox(0,0)[rt]{\lineheight{1.25}\smash{\begin{tabular}[t]{r}\texttt{\textbf{\footnotesize{54}}}\end{tabular}}}}%
    \put(0,0){\includegraphics[width=\unitlength,page=3]{plot-correlation.pdf}}%
    \put(0.07,0.24033333){\color[rgb]{0,0,0}\makebox(0,0)[rt]{\lineheight{1.25}\smash{\begin{tabular}[t]{r}\texttt{\textbf{\footnotesize{56}}}\end{tabular}}}}%
    \put(0,0){\includegraphics[width=\unitlength,page=4]{plot-correlation.pdf}}%
    \put(0.07,0.31983333){\color[rgb]{0,0,0}\makebox(0,0)[rt]{\lineheight{1.25}\smash{\begin{tabular}[t]{r}\texttt{\textbf{\footnotesize{58}}}\end{tabular}}}}%
    \put(0,0){\includegraphics[width=\unitlength,page=5]{plot-correlation.pdf}}%
    \put(0.07,0.39933333){\color[rgb]{0,0,0}\makebox(0,0)[rt]{\lineheight{1.25}\smash{\begin{tabular}[t]{r}\texttt{\textbf{\footnotesize{60}}}\end{tabular}}}}%
    \put(0,0){\includegraphics[width=\unitlength,page=6]{plot-correlation.pdf}}%
    \put(0.07,0.47883333){\color[rgb]{0,0,0}\makebox(0,0)[rt]{\lineheight{1.25}\smash{\begin{tabular}[t]{r}\texttt{\textbf{\footnotesize{62}}}\end{tabular}}}}%
    \put(0,0){\includegraphics[width=\unitlength,page=7]{plot-correlation.pdf}}%
    \put(0.07,0.55833333){\color[rgb]{0,0,0}\makebox(0,0)[rt]{\lineheight{1.25}\smash{\begin{tabular}[t]{r}\texttt{\textbf{\footnotesize{64}}}\end{tabular}}}}%
    \put(0,0){\includegraphics[width=\unitlength,page=8]{plot-correlation.pdf}}%
    \put(0.07,0.63783333){\color[rgb]{0,0,0}\makebox(0,0)[rt]{\lineheight{1.25}\smash{\begin{tabular}[t]{r}\texttt{\textbf{\footnotesize{66}}}\end{tabular}}}}%
    \put(0,0){\includegraphics[width=\unitlength,page=9]{plot-correlation.pdf}}%
    \put(0.08866667,0.04133333){\color[rgb]{0,0,0}\makebox(0,0)[t]{\lineheight{1.25}\smash{\begin{tabular}[t]{c}\texttt{\textbf{\footnotesize{52}}}\end{tabular}}}}%
    \put(0,0){\includegraphics[width=\unitlength,page=10]{plot-correlation.pdf}}%
    \put(0.21616667,0.04133333){\color[rgb]{0,0,0}\makebox(0,0)[t]{\lineheight{1.25}\smash{\begin{tabular}[t]{c}\texttt{\textbf{\footnotesize{54}}}\end{tabular}}}}%
    \put(0,0){\includegraphics[width=\unitlength,page=11]{plot-correlation.pdf}}%
    \put(0.34366667,0.04133333){\color[rgb]{0,0,0}\makebox(0,0)[t]{\lineheight{1.25}\smash{\begin{tabular}[t]{c}\texttt{\textbf{\footnotesize{56}}}\end{tabular}}}}%
    \put(0,0){\includegraphics[width=\unitlength,page=12]{plot-correlation.pdf}}%
    \put(0.47116667,0.04133333){\color[rgb]{0,0,0}\makebox(0,0)[t]{\lineheight{1.25}\smash{\begin{tabular}[t]{c}\texttt{\textbf{\footnotesize{58}}}\end{tabular}}}}%
    \put(0,0){\includegraphics[width=\unitlength,page=13]{plot-correlation.pdf}}%
    \put(0.59866667,0.04133333){\color[rgb]{0,0,0}\makebox(0,0)[t]{\lineheight{1.25}\smash{\begin{tabular}[t]{c}\texttt{\textbf{\footnotesize{60}}}\end{tabular}}}}%
    \put(0,0){\includegraphics[width=\unitlength,page=14]{plot-correlation.pdf}}%
    \put(0.72616667,0.04133333){\color[rgb]{0,0,0}\makebox(0,0)[t]{\lineheight{1.25}\smash{\begin{tabular}[t]{c}\texttt{\textbf{\footnotesize{62}}}\end{tabular}}}}%
    \put(0,0){\includegraphics[width=\unitlength,page=15]{plot-correlation.pdf}}%
    \put(0.85366667,0.04133333){\color[rgb]{0,0,0}\makebox(0,0)[t]{\lineheight{1.25}\smash{\begin{tabular}[t]{c}\texttt{\textbf{\footnotesize{64}}}\end{tabular}}}}%
    \put(0,0){\includegraphics[width=\unitlength,page=16]{plot-correlation.pdf}}%
    \put(0.98116667,0.04133333){\color[rgb]{0,0,0}\makebox(0,0)[t]{\lineheight{1.25}\smash{\begin{tabular}[t]{c}\texttt{\textbf{\footnotesize{66}}}\end{tabular}}}}%
    \put(0,0){\includegraphics[width=\unitlength,page=17]{plot-correlation.pdf}}%
    \put(0.026,0.36816667){\rotatebox{90}{\makebox(0,0)[t]{\lineheight{1.25}\smash{\begin{tabular}[t]{c}\texttt{\textbf{\footnotesize{\% of wins on fresh drafts}}}\end{tabular}}}}}%
    \put(0.53483333,0.01133333){\makebox(0,0)[t]{\lineheight{1.25}\smash{\begin{tabular}[t]{c}\texttt{\textbf{\footnotesize{\% of wins on known drafts}}}\end{tabular}}}}%
    \put(0.69983333,0.24133333){\makebox(0,0)[lt]{\lineheight{1.25}\smash{\begin{tabular}[t]{l}\texttt{\textbf{\footnotesize{AG}}}\end{tabular}}}}%
    \put(0,0){\includegraphics[width=\unitlength,page=18]{plot-correlation.pdf}}%
    \put(0.69983333,0.20133333){\makebox(0,0)[lt]{\lineheight{1.25}\smash{\begin{tabular}[t]{l}\texttt{\textbf{\footnotesize{AG\textsubscript{all}}}}\end{tabular}}}}%
    \put(0,0){\includegraphics[width=\unitlength,page=19]{plot-correlation.pdf}}%
    \put(0.69983333,0.16133333){\makebox(0,0)[lt]{\lineheight{1.25}\smash{\begin{tabular}[t]{l}\texttt{\textbf{\footnotesize{AG\textsubscript{weights}}}}\end{tabular}}}}%
    \put(0,0){\includegraphics[width=\unitlength,page=20]{plot-correlation.pdf}}%
    \put(0.69983333,0.12133333){\makebox(0,0)[lt]{\lineheight{1.25}\smash{\begin{tabular}[t]{l}\texttt{\textbf{\footnotesize{Evo\textsubscript{base}}}}\end{tabular}}}}%
    \put(0,0){\includegraphics[width=\unitlength,page=21]{plot-correlation.pdf}}%
  \end{picture}%
\endgroup%

%% file: plot-performance-random.pdf_tex
\begingroup%
  \makeatletter%
  \providecommand\color[2][]{%
    \errmessage{(Inkscape) Color is used for the text in Inkscape, but the package 'color.sty' is not loaded}%
    \renewcommand\color[2][]{}%
  }%
  \providecommand\transparent[1]{%
    \errmessage{(Inkscape) Transparency is used (non-zero) for the text in Inkscape, but the package 'transparent.sty' is not loaded}%
    \renewcommand\transparent[1]{}%
  }%
  \providecommand\rotatebox[2]{#2}%
  \newcommand*\fsize{\dimexpr\f@size pt\relax}%
  \newcommand*\lineheight[1]{\fontsize{\fsize}{#1\fsize}\selectfont}%
  \ifx\svgwidth\undefined%
    \setlength{\unitlength}{453.41689453bp}%
    \ifx\svgscale\undefined%
      \relax%
    \else%
      \setlength{\unitlength}{\unitlength * \real{\svgscale}}%
    \fi%
  \else%
    \setlength{\unitlength}{\svgwidth}%
  \fi%
  \global\let\svgwidth\undefined%
  \global\let\svgscale\undefined%
  \makeatother%
  \begin{picture}(1,0.66164275)%
    \lineheight{1}%
    \setlength\tabcolsep{0pt}%
    \put(0,0){\includegraphics[width=\unitlength,page=1]{plot-performance-random.pdf}}%
    \put(0.06245223,0.08138206){\makebox(0,0)[rt]{\lineheight{1.25}\smash{\begin{tabular}[t]{r}\texttt{\textbf{\footnotesize{51}}}\end{tabular}}}}%
    \put(0,0){\includegraphics[width=\unitlength,page=2]{plot-performance-random.pdf}}%
    \put(0.06245223,0.15912508){\makebox(0,0)[rt]{\lineheight{1.25}\smash{\begin{tabular}[t]{r}\texttt{\textbf{\footnotesize{52}}}\end{tabular}}}}%
    \put(0,0){\includegraphics[width=\unitlength,page=3]{plot-performance-random.pdf}}%
    \put(0.06245223,0.23670269){\makebox(0,0)[rt]{\lineheight{1.25}\smash{\begin{tabular}[t]{r}\texttt{\textbf{\footnotesize{53}}}\end{tabular}}}}%
    \put(0,0){\includegraphics[width=\unitlength,page=4]{plot-performance-random.pdf}}%
    \put(0.06245223,0.31444572){\makebox(0,0)[rt]{\lineheight{1.25}\smash{\begin{tabular}[t]{r}\texttt{\textbf{\footnotesize{54}}}\end{tabular}}}}%
    \put(0,0){\includegraphics[width=\unitlength,page=5]{plot-performance-random.pdf}}%
    \put(0.06245223,0.39218874){\makebox(0,0)[rt]{\lineheight{1.25}\smash{\begin{tabular}[t]{r}\texttt{\textbf{\footnotesize{55}}}\end{tabular}}}}%
    \put(0,0){\includegraphics[width=\unitlength,page=6]{plot-performance-random.pdf}}%
    \put(0.06245223,0.46993176){\makebox(0,0)[rt]{\lineheight{1.25}\smash{\begin{tabular}[t]{r}\texttt{\textbf{\footnotesize{56}}}\end{tabular}}}}%
    \put(0,0){\includegraphics[width=\unitlength,page=7]{plot-performance-random.pdf}}%
    \put(0.06245223,0.54750937){\makebox(0,0)[rt]{\lineheight{1.25}\smash{\begin{tabular}[t]{r}\texttt{\textbf{\footnotesize{57}}}\end{tabular}}}}%
    \put(0,0){\includegraphics[width=\unitlength,page=8]{plot-performance-random.pdf}}%
    \put(0.06245223,0.6252524){\makebox(0,0)[rt]{\lineheight{1.25}\smash{\begin{tabular}[t]{r}\texttt{\textbf{\footnotesize{58}}}\end{tabular}}}}%
    \put(0,0){\includegraphics[width=\unitlength,page=9]{plot-performance-random.pdf}}%
    \put(0.07618131,0.05160813){\makebox(0,0)[t]{\lineheight{1.25}\smash{\begin{tabular}[t]{c}\texttt{\textbf{\footnotesize{0}}}\end{tabular}}}}%
    \put(0,0){\includegraphics[width=\unitlength,page=10]{plot-performance-random.pdf}}%
    \put(0.25267452,0.05160813){\makebox(0,0)[t]{\lineheight{1.25}\smash{\begin{tabular}[t]{c}\texttt{\textbf{\footnotesize{2x10\textsuperscript{5}}}}\end{tabular}}}}%
    \put(0,0){\includegraphics[width=\unitlength,page=11]{plot-performance-random.pdf}}%
    \put(0.42916772,0.05160813){\makebox(0,0)[t]{\lineheight{1.25}\smash{\begin{tabular}[t]{c}\texttt{\textbf{\footnotesize{4x10\textsuperscript{5}}}}\end{tabular}}}}%
    \put(0,0){\includegraphics[width=\unitlength,page=12]{plot-performance-random.pdf}}%
    \put(0.60566092,0.05160813){\makebox(0,0)[t]{\lineheight{1.25}\smash{\begin{tabular}[t]{c}\texttt{\textbf{\footnotesize{6x10\textsuperscript{5}}}}\end{tabular}}}}%
    \put(0,0){\includegraphics[width=\unitlength,page=13]{plot-performance-random.pdf}}%
    \put(0.78215413,0.05160813){\makebox(0,0)[t]{\lineheight{1.25}\smash{\begin{tabular}[t]{c}\texttt{\textbf{\footnotesize{8x10\textsuperscript{5}}}}\end{tabular}}}}%
    \put(0,0){\includegraphics[width=\unitlength,page=14]{plot-performance-random.pdf}}%
    \put(0.95864733,0.05160813){\makebox(0,0)[t]{\lineheight{1.25}\smash{\begin{tabular}[t]{c}\texttt{\textbf{\footnotesize{10x10\textsuperscript{5}}}}\end{tabular}}}}%
    \put(0,0){\includegraphics[width=\unitlength,page=15]{plot-performance-random.pdf}}%
    \put(0.01613723,0.35976824){\rotatebox{90}{\makebox(0,0)[t]{\lineheight{1.25}\smash{\begin{tabular}[t]{c}\texttt{\textbf{\footnotesize{\% of wins}}}\end{tabular}}}}}%
    \put(0.51733162,0.01439073){\makebox(0,0)[t]{\lineheight{1.25}\smash{\begin{tabular}[t]{c}\texttt{\textbf{\footnotesize{Computation cost (games)}}}\end{tabular}}}}%
    \put(0.42007013,0.14092991){\makebox(0,0)[rt]{\lineheight{1.25}\smash{\begin{tabular}[t]{r}\texttt{\textbf{\footnotesize{AG}}}\end{tabular}}}}%
    \put(0,0){\includegraphics[width=\unitlength,page=16]{plot-performance-random.pdf}}%
    \put(0.42007013,0.11115598){\makebox(0,0)[rt]{\lineheight{1.25}\smash{\begin{tabular}[t]{r}\texttt{\textbf{\footnotesize{AG\textsubscript{weights}}}}\end{tabular}}}}%
    \put(0,0){\includegraphics[width=\unitlength,page=17]{plot-performance-random.pdf}}%
    \put(0.65462249,0.14092991){\makebox(0,0)[rt]{\lineheight{1.25}\smash{\begin{tabular}[t]{r}\texttt{\textbf{\footnotesize{AG\textsubscript{all}}}}\end{tabular}}}}%
    \put(0,0){\includegraphics[width=\unitlength,page=18]{plot-performance-random.pdf}}%
    \put(0.65462249,0.11115598){\makebox(0,0)[rt]{\lineheight{1.25}\smash{\begin{tabular}[t]{r}\texttt{\textbf{\footnotesize{Evo\textsubscript{base}}}}\end{tabular}}}}%
    \put(0,0){\includegraphics[width=\unitlength,page=19]{plot-performance-random.pdf}}%
  \end{picture}%
\endgroup%

%% file: plot-performance-greedy.pdf_tex
\begingroup%
  \makeatletter%
  \providecommand\color[2][]{%
    \errmessage{(Inkscape) Color is used for the text in Inkscape, but the package 'color.sty' is not loaded}%
    \renewcommand\color[2][]{}%
  }%
  \providecommand\transparent[1]{%
    \errmessage{(Inkscape) Transparency is used (non-zero) for the text in Inkscape, but the package 'transparent.sty' is not loaded}%
    \renewcommand\transparent[1]{}%
  }%
  \providecommand\rotatebox[2]{#2}%
  \newcommand*\fsize{\dimexpr\f@size pt\relax}%
  \newcommand*\lineheight[1]{\fontsize{\fsize}{#1\fsize}\selectfont}%
  \ifx\svgwidth\undefined%
    \setlength{\unitlength}{453.41689453bp}%
    \ifx\svgscale\undefined%
      \relax%
    \else%
      \setlength{\unitlength}{\unitlength * \real{\svgscale}}%
    \fi%
  \else%
    \setlength{\unitlength}{\svgwidth}%
  \fi%
  \global\let\svgwidth\undefined%
  \global\let\svgscale\undefined%
  \makeatother%
  \begin{picture}(1,0.66164275)%
    \lineheight{1}%
    \setlength\tabcolsep{0pt}%
    \put(0,0){\includegraphics[width=\unitlength,page=1]{plot-performance-greedy.pdf}}%
    \put(0.06245223,0.08138206){\makebox(0,0)[rt]{\lineheight{1.25}\smash{\begin{tabular}[t]{r}\texttt{\textbf{\footnotesize{60}}}\end{tabular}}}}%
    \put(0,0){\includegraphics[width=\unitlength,page=2]{plot-performance-greedy.pdf}}%
    \put(0.06245223,0.15912508){\makebox(0,0)[rt]{\lineheight{1.25}\smash{\begin{tabular}[t]{r}\texttt{\textbf{\footnotesize{62}}}\end{tabular}}}}%
    \put(0,0){\includegraphics[width=\unitlength,page=3]{plot-performance-greedy.pdf}}%
    \put(0.06245223,0.23670269){\makebox(0,0)[rt]{\lineheight{1.25}\smash{\begin{tabular}[t]{r}\texttt{\textbf{\footnotesize{64}}}\end{tabular}}}}%
    \put(0,0){\includegraphics[width=\unitlength,page=4]{plot-performance-greedy.pdf}}%
    \put(0.06245223,0.31444572){\makebox(0,0)[rt]{\lineheight{1.25}\smash{\begin{tabular}[t]{r}\texttt{\textbf{\footnotesize{66}}}\end{tabular}}}}%
    \put(0,0){\includegraphics[width=\unitlength,page=5]{plot-performance-greedy.pdf}}%
    \put(0.06245223,0.39218874){\makebox(0,0)[rt]{\lineheight{1.25}\smash{\begin{tabular}[t]{r}\texttt{\textbf{\footnotesize{68}}}\end{tabular}}}}%
    \put(0,0){\includegraphics[width=\unitlength,page=6]{plot-performance-greedy.pdf}}%
    \put(0.06245223,0.46993176){\makebox(0,0)[rt]{\lineheight{1.25}\smash{\begin{tabular}[t]{r}\texttt{\textbf{\footnotesize{70}}}\end{tabular}}}}%
    \put(0,0){\includegraphics[width=\unitlength,page=7]{plot-performance-greedy.pdf}}%
    \put(0.06245223,0.54750937){\makebox(0,0)[rt]{\lineheight{1.25}\smash{\begin{tabular}[t]{r}\texttt{\textbf{\footnotesize{72}}}\end{tabular}}}}%
    \put(0,0){\includegraphics[width=\unitlength,page=8]{plot-performance-greedy.pdf}}%
    \put(0.06245223,0.6252524){\makebox(0,0)[rt]{\lineheight{1.25}\smash{\begin{tabular}[t]{r}\texttt{\textbf{\footnotesize{74}}}\end{tabular}}}}%
    \put(0,0){\includegraphics[width=\unitlength,page=9]{plot-performance-greedy.pdf}}%
    \put(0.07618131,0.05160813){\makebox(0,0)[t]{\lineheight{1.25}\smash{\begin{tabular}[t]{c}\texttt{\textbf{\footnotesize{0}}}\end{tabular}}}}%
    \put(0,0){\includegraphics[width=\unitlength,page=10]{plot-performance-greedy.pdf}}%
    \put(0.25267452,0.05160813){\makebox(0,0)[t]{\lineheight{1.25}\smash{\begin{tabular}[t]{c}\texttt{\textbf{\footnotesize{2x10\textsuperscript{5}}}}\end{tabular}}}}%
    \put(0,0){\includegraphics[width=\unitlength,page=11]{plot-performance-greedy.pdf}}%
    \put(0.42916772,0.05160813){\makebox(0,0)[t]{\lineheight{1.25}\smash{\begin{tabular}[t]{c}\texttt{\textbf{\footnotesize{4x10\textsuperscript{5}}}}\end{tabular}}}}%
    \put(0,0){\includegraphics[width=\unitlength,page=12]{plot-performance-greedy.pdf}}%
    \put(0.60566092,0.05160813){\makebox(0,0)[t]{\lineheight{1.25}\smash{\begin{tabular}[t]{c}\texttt{\textbf{\footnotesize{6x10\textsuperscript{5}}}}\end{tabular}}}}%
    \put(0,0){\includegraphics[width=\unitlength,page=13]{plot-performance-greedy.pdf}}%
    \put(0.78215413,0.05160813){\makebox(0,0)[t]{\lineheight{1.25}\smash{\begin{tabular}[t]{c}\texttt{\textbf{\footnotesize{8x10\textsuperscript{5}}}}\end{tabular}}}}%
    \put(0,0){\includegraphics[width=\unitlength,page=14]{plot-performance-greedy.pdf}}%
    \put(0.95864733,0.05160813){\makebox(0,0)[t]{\lineheight{1.25}\smash{\begin{tabular}[t]{c}\texttt{\textbf{\footnotesize{10x10\textsuperscript{5}}}}\end{tabular}}}}%
    \put(0,0){\includegraphics[width=\unitlength,page=15]{plot-performance-greedy.pdf}}%
    \put(0.01613723,0.35976824){\rotatebox{90}{\makebox(0,0)[t]{\lineheight{1.25}\smash{\begin{tabular}[t]{c}\texttt{\textbf{\footnotesize{\% of wins}}}\end{tabular}}}}}%
    \put(0.51733162,0.01439073){\makebox(0,0)[t]{\lineheight{1.25}\smash{\begin{tabular}[t]{c}\texttt{\textbf{\footnotesize{Computation cost (games)}}}\end{tabular}}}}%
    \put(0.42007013,0.14092991){\makebox(0,0)[rt]{\lineheight{1.25}\smash{\begin{tabular}[t]{r}\texttt{\textbf{\footnotesize{AG}}}\end{tabular}}}}%
    \put(0,0){\includegraphics[width=\unitlength,page=16]{plot-performance-greedy.pdf}}%
    \put(0.42007013,0.11115598){\makebox(0,0)[rt]{\lineheight{1.25}\smash{\begin{tabular}[t]{r}\texttt{\textbf{\footnotesize{AG\textsubscript{weights}}}}\end{tabular}}}}%
    \put(0,0){\includegraphics[width=\unitlength,page=17]{plot-performance-greedy.pdf}}%
    \put(0.65462249,0.14092991){\makebox(0,0)[rt]{\lineheight{1.25}\smash{\begin{tabular}[t]{r}\texttt{\textbf{\footnotesize{AG\textsubscript{all}}}}\end{tabular}}}}%
    \put(0,0){\includegraphics[width=\unitlength,page=18]{plot-performance-greedy.pdf}}%
    \put(0.65462249,0.11115598){\makebox(0,0)[rt]{\lineheight{1.25}\smash{\begin{tabular}[t]{r}\texttt{\textbf{\footnotesize{Evo\textsubscript{base}}}}\end{tabular}}}}%
    \put(0,0){\includegraphics[width=\unitlength,page=19]{plot-performance-greedy.pdf}}%
  \end{picture}%
\endgroup%

%% file: plot-champions.pdf_tex
\begingroup%
  \makeatletter%
  \providecommand\color[2][]{%
    \errmessage{(Inkscape) Color is used for the text in Inkscape, but the package 'color.sty' is not loaded}%
    \renewcommand\color[2][]{}%
  }%
  \providecommand\transparent[1]{%
    \errmessage{(Inkscape) Transparency is used (non-zero) for the text in Inkscape, but the package 'transparent.sty' is not loaded}%
    \renewcommand\transparent[1]{}%
  }%
  \providecommand\rotatebox[2]{#2}%
  \newcommand*\fsize{\dimexpr\f@size pt\relax}%
  \newcommand*\lineheight[1]{\fontsize{\fsize}{#1\fsize}\selectfont}%
  \ifx\svgwidth\undefined%
    \setlength{\unitlength}{750bp}%
    \ifx\svgscale\undefined%
      \relax%
    \else%
      \setlength{\unitlength}{\unitlength * \real{\svgscale}}%
    \fi%
  \else%
    \setlength{\unitlength}{\svgwidth}%
  \fi%
  \global\let\svgwidth\undefined%
  \global\let\svgscale\undefined%
  \makeatother%
  \begin{picture}(1,0.6)%
    \lineheight{1}%
    \setlength\tabcolsep{0pt}%
    \put(0,0){\includegraphics[width=\unitlength,page=1]{plot-champions.pdf}}%
    \put(0.0448,0.0596){\makebox(0,0)[rt]{\lineheight{1.25}\smash{\begin{tabular}[t]{r}\texttt{\textbf{\small{48}}}\end{tabular}}}}%
    \put(0,0){\includegraphics[width=\unitlength,page=2]{plot-champions.pdf}}%
    \put(0.0448,0.1448){\makebox(0,0)[rt]{\lineheight{1.25}\smash{\begin{tabular}[t]{r}\texttt{\textbf{\small{50}}}\end{tabular}}}}%
    \put(0,0){\includegraphics[width=\unitlength,page=3]{plot-champions.pdf}}%
    \put(0.0448,0.23){\makebox(0,0)[rt]{\lineheight{1.25}\smash{\begin{tabular}[t]{r}\texttt{\textbf{\small{52}}}\end{tabular}}}}%
    \put(0,0){\includegraphics[width=\unitlength,page=4]{plot-champions.pdf}}%
    \put(0.0448,0.3152){\makebox(0,0)[rt]{\lineheight{1.25}\smash{\begin{tabular}[t]{r}\texttt{\textbf{\small{54}}}\end{tabular}}}}%
    \put(0,0){\includegraphics[width=\unitlength,page=5]{plot-champions.pdf}}%
    \put(0.0448,0.4003){\makebox(0,0)[rt]{\lineheight{1.25}\smash{\begin{tabular}[t]{r}\texttt{\textbf{\small{56}}}\end{tabular}}}}%
    \put(0,0){\includegraphics[width=\unitlength,page=6]{plot-champions.pdf}}%
    \put(0.0448,0.4855){\makebox(0,0)[rt]{\lineheight{1.25}\smash{\begin{tabular}[t]{r}\texttt{\textbf{\small{58}}}\end{tabular}}}}%
    \put(0,0){\includegraphics[width=\unitlength,page=7]{plot-champions.pdf}}%
    \put(0.0448,0.5707){\makebox(0,0)[rt]{\lineheight{1.25}\smash{\begin{tabular}[t]{r}\texttt{\textbf{\small{60}}}\end{tabular}}}}%
    \put(0,0){\includegraphics[width=\unitlength,page=8]{plot-champions.pdf}}%
    \put(0.056,0.0356){\makebox(0,0)[t]{\lineheight{1.25}\smash{\begin{tabular}[t]{c}\texttt{\textbf{\small{0}}}\end{tabular}}}}%
    \put(0,0){\includegraphics[width=\unitlength,page=9]{plot-champions.pdf}}%
    \put(0.147,0.0356){\makebox(0,0)[t]{\lineheight{1.25}\smash{\begin{tabular}[t]{c}\texttt{\textbf{\small{100}}}\end{tabular}}}}%
    \put(0,0){\includegraphics[width=\unitlength,page=10]{plot-champions.pdf}}%
    \put(0.2381,0.0356){\makebox(0,0)[t]{\lineheight{1.25}\smash{\begin{tabular}[t]{c}\texttt{\textbf{\small{200}}}\end{tabular}}}}%
    \put(0,0){\includegraphics[width=\unitlength,page=11]{plot-champions.pdf}}%
    \put(0.3291,0.0356){\makebox(0,0)[t]{\lineheight{1.25}\smash{\begin{tabular}[t]{c}\texttt{\textbf{\small{300}}}\end{tabular}}}}%
    \put(0,0){\includegraphics[width=\unitlength,page=12]{plot-champions.pdf}}%
    \put(0.4201,0.0356){\makebox(0,0)[t]{\lineheight{1.25}\smash{\begin{tabular}[t]{c}\texttt{\textbf{\small{400}}}\end{tabular}}}}%
    \put(0,0){\includegraphics[width=\unitlength,page=13]{plot-champions.pdf}}%
    \put(0.5112,0.0356){\makebox(0,0)[t]{\lineheight{1.25}\smash{\begin{tabular}[t]{c}\texttt{\textbf{\small{500}}}\end{tabular}}}}%
    \put(0,0){\includegraphics[width=\unitlength,page=14]{plot-champions.pdf}}%
    \put(0.6022,0.0356){\makebox(0,0)[t]{\lineheight{1.25}\smash{\begin{tabular}[t]{c}\texttt{\textbf{\small{600}}}\end{tabular}}}}%
    \put(0,0){\includegraphics[width=\unitlength,page=15]{plot-champions.pdf}}%
    \put(0.6932,0.0356){\makebox(0,0)[t]{\lineheight{1.25}\smash{\begin{tabular}[t]{c}\texttt{\textbf{\small{700}}}\end{tabular}}}}%
    \put(0,0){\includegraphics[width=\unitlength,page=16]{plot-champions.pdf}}%
    \put(0.7842,0.0356){\makebox(0,0)[t]{\lineheight{1.25}\smash{\begin{tabular}[t]{c}\texttt{\textbf{\small{800}}}\end{tabular}}}}%
    \put(0,0){\includegraphics[width=\unitlength,page=17]{plot-champions.pdf}}%
    \put(0.8753,0.0356){\makebox(0,0)[t]{\lineheight{1.25}\smash{\begin{tabular}[t]{c}\texttt{\textbf{\small{900}}}\end{tabular}}}}%
    \put(0,0){\includegraphics[width=\unitlength,page=18]{plot-champions.pdf}}%
    \put(0.9663,0.0356){\makebox(0,0)[t]{\lineheight{1.25}\smash{\begin{tabular}[t]{c}\texttt{\textbf{\small{1000}}}\end{tabular}}}}%
    \put(0,0){\includegraphics[width=\unitlength,page=19]{plot-champions.pdf}}%
    \put(0,0.3203){\rotatebox{90}{\makebox(0,0)[t]{\lineheight{1.25}\smash{\begin{tabular}[t]{c}\texttt{\textbf{\small{\% of wins}}}\end{tabular}}}}}%
    \put(0.5111,0){\makebox(0,0)[t]{\lineheight{1.25}\smash{\begin{tabular}[t]{c}\texttt{\textbf{\small{Generation}}}\end{tabular}}}}%
    \put(0.8759,0.5467){\makebox(0,0)[rt]{\lineheight{1.25}\smash{\begin{tabular}[t]{r}\texttt{\textbf{\small{Champions\ \ \ \ 0 (avg.\ 49.44)}}}\end{tabular}}}}%
    \put(0,0){\includegraphics[width=\unitlength,page=20]{plot-champions.pdf}}%
    \put(0.8759,0.5227){\makebox(0,0)[rt]{\lineheight{1.25}\smash{\begin{tabular}[t]{r}\texttt{\textbf{\small{Champions\ \ 200 (avg.\ 52.81)}}}\end{tabular}}}}%
    \put(0,0){\includegraphics[width=\unitlength,page=21]{plot-champions.pdf}}%
    \put(0.8759,0.4987){\makebox(0,0)[rt]{\lineheight{1.25}\smash{\begin{tabular}[t]{r}\texttt{\textbf{\small{Champions\ \ 400 (avg.\ 53.06)}}}\end{tabular}}}}%
    \put(0,0){\includegraphics[width=\unitlength,page=22]{plot-champions.pdf}}%
    \put(0.8759,0.4747){\makebox(0,0)[rt]{\lineheight{1.25}\smash{\begin{tabular}[t]{r}\texttt{\textbf{\small{Champions\ \ 600 (avg.\ 53.47)}}}\end{tabular}}}}%
    \put(0,0){\includegraphics[width=\unitlength,page=23]{plot-champions.pdf}}%
    \put(0.8759,0.4507){\makebox(0,0)[rt]{\lineheight{1.25}\smash{\begin{tabular}[t]{r}\texttt{\textbf{\small{Champions\ \ 800 (avg.\ 52.82)}}}\end{tabular}}}}%
    \put(0,0){\includegraphics[width=\unitlength,page=24]{plot-champions.pdf}}%
    \put(0.8759,0.4267){\makebox(0,0)[rt]{\lineheight{1.25}\smash{\begin{tabular}[t]{r}\texttt{\textbf{\small{Champions\ 1000 (avg.\ 54.24)}}}\end{tabular}}}}%
    \put(0,0){\includegraphics[width=\unitlength,page=25]{plot-champions.pdf}}%
  \end{picture}%
\endgroup%

%% file: plot-performance-random-variants.pdf_tex
\begingroup%
  \makeatletter%
  \providecommand\color[2][]{%
    \errmessage{(Inkscape) Color is used for the text in Inkscape, but the package 'color.sty' is not loaded}%
    \renewcommand\color[2][]{}%
  }%
  \providecommand\transparent[1]{%
    \errmessage{(Inkscape) Transparency is used (non-zero) for the text in Inkscape, but the package 'transparent.sty' is not loaded}%
    \renewcommand\transparent[1]{}%
  }%
  \providecommand\rotatebox[2]{#2}%
  \newcommand*\fsize{\dimexpr\f@size pt\relax}%
  \newcommand*\lineheight[1]{\fontsize{\fsize}{#1\fsize}\selectfont}%
  \ifx\svgwidth\undefined%
    \setlength{\unitlength}{453.41689453bp}%
    \ifx\svgscale\undefined%
      \relax%
    \else%
      \setlength{\unitlength}{\unitlength * \real{\svgscale}}%
    \fi%
  \else%
    \setlength{\unitlength}{\svgwidth}%
  \fi%
  \global\let\svgwidth\undefined%
  \global\let\svgscale\undefined%
  \makeatother%
  \begin{picture}(1,0.66164275)%
    \lineheight{1}%
    \setlength\tabcolsep{0pt}%
    \put(0,0){\includegraphics[width=\unitlength,page=1]{plot-performance-random-variants.pdf}}%
    \put(0.06245223,0.08138206){\makebox(0,0)[rt]{\lineheight{1.25}\smash{\begin{tabular}[t]{r}\texttt{\textbf{\footnotesize{51}}}\end{tabular}}}}%
    \put(0,0){\includegraphics[width=\unitlength,page=2]{plot-performance-random-variants.pdf}}%
    \put(0.06245223,0.15912508){\makebox(0,0)[rt]{\lineheight{1.25}\smash{\begin{tabular}[t]{r}\texttt{\textbf{\footnotesize{52}}}\end{tabular}}}}%
    \put(0,0){\includegraphics[width=\unitlength,page=3]{plot-performance-random-variants.pdf}}%
    \put(0.06245223,0.23670269){\makebox(0,0)[rt]{\lineheight{1.25}\smash{\begin{tabular}[t]{r}\texttt{\textbf{\footnotesize{53}}}\end{tabular}}}}%
    \put(0,0){\includegraphics[width=\unitlength,page=4]{plot-performance-random-variants.pdf}}%
    \put(0.06245223,0.31444572){\makebox(0,0)[rt]{\lineheight{1.25}\smash{\begin{tabular}[t]{r}\texttt{\textbf{\footnotesize{54}}}\end{tabular}}}}%
    \put(0,0){\includegraphics[width=\unitlength,page=5]{plot-performance-random-variants.pdf}}%
    \put(0.06245223,0.39218874){\makebox(0,0)[rt]{\lineheight{1.25}\smash{\begin{tabular}[t]{r}\texttt{\textbf{\footnotesize{55}}}\end{tabular}}}}%
    \put(0,0){\includegraphics[width=\unitlength,page=6]{plot-performance-random-variants.pdf}}%
    \put(0.06245223,0.46993176){\makebox(0,0)[rt]{\lineheight{1.25}\smash{\begin{tabular}[t]{r}\texttt{\textbf{\footnotesize{56}}}\end{tabular}}}}%
    \put(0,0){\includegraphics[width=\unitlength,page=7]{plot-performance-random-variants.pdf}}%
    \put(0.06245223,0.54750937){\makebox(0,0)[rt]{\lineheight{1.25}\smash{\begin{tabular}[t]{r}\texttt{\textbf{\footnotesize{57}}}\end{tabular}}}}%
    \put(0,0){\includegraphics[width=\unitlength,page=8]{plot-performance-random-variants.pdf}}%
    \put(0.06245223,0.6252524){\makebox(0,0)[rt]{\lineheight{1.25}\smash{\begin{tabular}[t]{r}\texttt{\textbf{\footnotesize{58}}}\end{tabular}}}}%
    \put(0,0){\includegraphics[width=\unitlength,page=9]{plot-performance-random-variants.pdf}}%
    \put(0.07618131,0.05160813){\makebox(0,0)[t]{\lineheight{1.25}\smash{\begin{tabular}[t]{c}\texttt{\textbf{\footnotesize{0}}}\end{tabular}}}}%
    \put(0,0){\includegraphics[width=\unitlength,page=10]{plot-performance-random-variants.pdf}}%
    \put(0.25267452,0.05160813){\makebox(0,0)[t]{\lineheight{1.25}\smash{\begin{tabular}[t]{c}\texttt{\textbf{\footnotesize{2x10\textsuperscript{5}}}}\end{tabular}}}}%
    \put(0,0){\includegraphics[width=\unitlength,page=11]{plot-performance-random-variants.pdf}}%
    \put(0.42916772,0.05160813){\makebox(0,0)[t]{\lineheight{1.25}\smash{\begin{tabular}[t]{c}\texttt{\textbf{\footnotesize{4x10\textsuperscript{5}}}}\end{tabular}}}}%
    \put(0,0){\includegraphics[width=\unitlength,page=12]{plot-performance-random-variants.pdf}}%
    \put(0.60566092,0.05160813){\makebox(0,0)[t]{\lineheight{1.25}\smash{\begin{tabular}[t]{c}\texttt{\textbf{\footnotesize{6x10\textsuperscript{5}}}}\end{tabular}}}}%
    \put(0,0){\includegraphics[width=\unitlength,page=13]{plot-performance-random-variants.pdf}}%
    \put(0.78215413,0.05160813){\makebox(0,0)[t]{\lineheight{1.25}\smash{\begin{tabular}[t]{c}\texttt{\textbf{\footnotesize{8x10\textsuperscript{5}}}}\end{tabular}}}}%
    \put(0,0){\includegraphics[width=\unitlength,page=14]{plot-performance-random-variants.pdf}}%
    \put(0.95864733,0.05160813){\makebox(0,0)[t]{\lineheight{1.25}\smash{\begin{tabular}[t]{c}\texttt{\textbf{\footnotesize{10x10\textsuperscript{5}}}}\end{tabular}}}}%
    \put(0,0){\includegraphics[width=\unitlength,page=15]{plot-performance-random-variants.pdf}}%
    \put(0.01613723,0.35976824){\rotatebox{90}{\makebox(0,0)[t]{\lineheight{1.25}\smash{\begin{tabular}[t]{c}\texttt{\textbf{\footnotesize{\% of wins}}}\end{tabular}}}}}%
    \put(0.51733162,0.01439073){\makebox(0,0)[t]{\lineheight{1.25}\smash{\begin{tabular}[t]{c}\texttt{\textbf{\footnotesize{Computation cost (games)}}}\end{tabular}}}}%
    \put(0.42007013,0.14092991){\makebox(0,0)[rt]{\lineheight{1.25}\smash{\begin{tabular}[t]{r}\texttt{\textbf{\footnotesize{AG\textsubscript{weights/2d}}}}\end{tabular}}}}%
    \put(0,0){\includegraphics[width=\unitlength,page=16]{plot-performance-random-variants.pdf}}%
    \put(0.42007013,0.11115598){\makebox(0,0)[rt]{\lineheight{1.25}\smash{\begin{tabular}[t]{r}\texttt{\textbf{\footnotesize{AG\textsubscript{weights/4d}}}}\end{tabular}}}}%
    \put(0,0){\includegraphics[width=\unitlength,page=17]{plot-performance-random-variants.pdf}}%
    \put(0.69580975,0.14092991){\makebox(0,0)[rt]{\lineheight{1.25}\smash{\begin{tabular}[t]{r}\texttt{\textbf{\footnotesize{AG\textsubscript{weights/2g}}}}\end{tabular}}}}%
    \put(0,0){\includegraphics[width=\unitlength,page=18]{plot-performance-random-variants.pdf}}%
    \put(0.69580975,0.11115598){\makebox(0,0)[rt]{\lineheight{1.25}\smash{\begin{tabular}[t]{r}\texttt{\textbf{\footnotesize{AG\textsubscript{weights/4g}}}}\end{tabular}}}}%
    \put(0,0){\includegraphics[width=\unitlength,page=19]{plot-performance-random-variants.pdf}}%
  \end{picture}%
\endgroup%

%% file: paper.bbl
\begin{thebibliography}{10}
\providecommand{\url}[1]{#1}
\csname url@samestyle\endcsname
\providecommand{\newblock}{\relax}
\providecommand{\bibinfo}[2]{#2}
\providecommand{\BIBentrySTDinterwordspacing}{\spaceskip=0pt\relax}
\providecommand{\BIBentryALTinterwordstretchfactor}{4}
\providecommand{\BIBentryALTinterwordspacing}{\spaceskip=\fontdimen2\font plus
\BIBentryALTinterwordstretchfactor\fontdimen3\font minus
  \fontdimen4\font\relax}
\providecommand{\BIBforeignlanguage}[2]{{%
\expandafter\ifx\csname l@#1\endcsname\relax
\typeout{** WARNING: IEEEtran.bst: No hyphenation pattern has been}%
\typeout{** loaded for the language `#1'. Using the pattern for}%
\typeout{** the default language instead.}%
\else
\language=\csname l@#1\endcsname
\fi
#2}}
\providecommand{\BIBdecl}{\relax}
\BIBdecl

\bibitem{Campbell2002Deep}
M.~Campbell, A.~J. Hoane, and F.~Hsu, ``{Deep Blue},'' \emph{Artificial
  intelligence}, vol. 134, no.~1, pp. 57--83, 2002.

\bibitem{Silver2016Mastering}
D.~Silver, A.~Huang, C.~J. Maddison, A.~Guez, L.~Sifre, G.~van~den Driessche,
  J.~Schrittwieser, I.~Antonoglou, V.~Panneershelvam, M.~Lanctot, S.~Dieleman,
  D.~Grewe, J.~Nham, N.~Kalchbrenner, I.~Sutskever, T.~Lillicrap, M.~Leach,
  K.~Kavukcuoglu, T.~Graepel, and D.~Hassabis, ``{Mastering the game of Go with
  deep neural networks and tree search},'' \emph{Nature}, vol. 529, pp.
  484--503, 2016.

\bibitem{OpenAIDota}
OpenAI, ``{OpenAI Five},'' \url{https://blog.openai.com/openai-five/}, 2017.

\bibitem{vinyals2019grandmaster}
O.~Vinyals, I.~Babuschkin, W.~M. Czarnecki, M.~Mathieu, A.~Dudzik, J.~Chung,
  D.~H. Choi, R.~Powell, T.~Ewalds, P.~Georgiev \emph{et~al.}, ``{Grandmaster
  level in StarCraft II using multi-agent reinforcement learning},''
  \emph{Nature}, vol. 575, no. 7782, pp. 350--354, 2019.

\bibitem{hoover2019many}
A.~K. Hoover, J.~Togelius, S.~Lee, and F.~de~Mesentier~Silva, ``{The Many AI
  Challenges of Hearthstone},'' \emph{KI-K{\"u}nstliche Intelligenz}, pp.
  1--11, 2019.

\bibitem{HearthstoneAICompetition}
A.~Dockhorn and S.~Mostaghim, ``{Hearthstone AI Competition},''
  \url{https://dockhorn.antares.uberspace.de/wordpress/}, 2018.

\bibitem{Blizzard2004Hearthstone}
{Blizzard Entertainment}, \emph{{Hearthstone}}, Blizzard Entertainment, 2004.

\bibitem{LOCMPage}
J.~Kowalski and R.~Miernik, ``{Legends of Code and Magic},''
  \url{http://legendsofcodeandmagic.com}, 2018.

\bibitem{janusz2017helping}
A.~Janusz, T.~Tajmajer, and M.~{\'S}wiechowski, ``{Helping AI to Play
  Hearthstone: AAIA'17 Data Mining Challenge},'' in \emph{2017 Federated
  Conference on Computer Science and Information Systems}.\hskip 1em plus 0.5em
  minus 0.4em\relax IEEE, 2017, pp. 121--125.

\bibitem{WotC1993MtG}
{R. Garfield}, \emph{{Magic: the Gathering}}, Wizards of the Coast, 1993.

\bibitem{Bethesda2017TESL}
{Dire Wolf Digital and Sparkypants Studios}, \emph{{The Elder Scrolls:
  Legends}}, Bethesda Softworks, 2017.

\bibitem{garcia2016evolutionary}
P.~Garc{\'\i}a-S{\'a}nchez, A.~Tonda, G.~Squillero, A.~Mora, and J.~J. Merelo,
  ``{Evolutionary deckbuilding in Hearthstone},'' in \emph{Computational
  Intelligence and Games (CIG), 2016 IEEE Conference on}, 2016, pp. 1--8.

\bibitem{bjorke2017deckbuilding}
S.~J. Bj{\o}rke and K.~A. Fludal, ``Deckbuilding in magic: The gathering using
  a genetic algorithm,'' Master's thesis, NTNU, 2017.

\bibitem{chen2018q}
Z.~Chen, C.~Amato, T.-H.~D. Nguyen, S.~Cooper, Y.~Sun, and M.~S. El-Nasr,
  ``{Q-deckrec: A fast deck recommendation system for collectible card
  games},'' in \emph{2018 IEEE Conference on Computational Intelligence and
  Games (CIG)}.\hskip 1em plus 0.5em minus 0.4em\relax IEEE, 2018, pp. 1--8.

\bibitem{bhatt2018exploring}
A.~Bhatt, S.~Lee, F.~de~Mesentier~Silva, C.~W. Watson, J.~Togelius, and A.~K.
  Hoover, ``{Exploring the Hearthstone deck space},'' in \emph{Proceedings of
  the 13th International Conference on the Foundations of Digital Games}, 2018,
  pp. 1--10.

\bibitem{volz2016demonstrating}
V.~Volz, G.~Rudolph, and B.~Naujoks, ``Demonstrating the feasibility of
  automatic game balancing,'' in \emph{Proceedings of the Genetic and
  Evolutionary Computation Conference}.\hskip 1em plus 0.5em minus 0.4em\relax
  ACM, 2016, pp. 269--276.

\bibitem{Silva2019evolving}
F.~de~Mesentier~Silva, R.~Canaan, S.~Lee, M.~C. Fontaine, J.~Togelius, and
  A.~K. Hoover, ``{Evolving the Hearthstone meta},'' in \emph{2019 IEEE
  Conference on Games (CoG)}, 2019, pp. 1--8.

\bibitem{Fontaine2019Mapping}
M.~C. Fontaine, S.~Lee, L.~B. Soros, F.~De~Mesentier~Silva, J.~Togelius, and
  A.~K. Hoover, ``{Mapping Hearthstone Deck Spaces Through MAP-elites with
  Sliding Boundaries},'' in \emph{Proceedings of the Genetic and Evolutionary
  Computation Conference}, 2019, pp. 161--169.

\bibitem{Lightforge2016}
ADWCTA and MERPS, ``{Lightforge: Hearthstone Arena Tier List},''
  http://thelightforge.com/TierList, 2016.

\bibitem{HearthArena2017}
HearthArena, ``{HearthArena: Beyond the Tier List },''
  https://www.heartharena.com/tierlist, 2017.

\bibitem{LOCMCG}
CodinGame, ``{Legends of Code and Magic -- Multiplayer Game},''
  \url{https://www.codingame.com/multiplayer/bot-programming/legends-of-code-magic},
  2018.

\bibitem{CG2018LOCMPostmortem}
------, ``{Legends of Code \& Magic (CC05) - Feedback {\&} Strategies},''
  https://www.codingame.com/forum/t/legends-of-code-magic-cc05-feedback-strategies/,
  2018.

\bibitem{ActiveGenesRepo}
J.~Kowalski and R.~Miernik, ``{Active Genes Evolution for LOCM -- source
  code},''
  \url{https://github.com/acatai/Strategy-Card-Game-AI-Competition/tree/master/paper-activegenes},
  2020.

\bibitem{LOCMSource}
------, ``{Strategy Card Game AI Competition -- source code},''
  \url{https://github.com/acatai/Strategy-Card-Game-AI-Competition}, 2019.

\end{thebibliography}
